\documentclass{article}
\usepackage{iclr2027_conference,times}

\iclrfinalcopy  

\usepackage{amsmath,amsfonts,bm}

\def\eqref#1{equation~\ref{#1}}

\def\1{\bm{1}}

\DeclareMathAlphabet{\mathsfit}{\encodingdefault}{\sfdefault}{m}{sl}
\SetMathAlphabet{\mathsfit}{bold}{\encodingdefault}{\sfdefault}{bx}{n}

\usepackage{graphicx}
\usepackage{booktabs}
\usepackage{colortbl}
\usepackage{arydshln}

\usepackage{comment}
\usepackage{subcaption}
\usepackage{enumitem}

\usepackage{caption}
\usepackage{multirow}

\usepackage{float}

\usepackage{array,tabularx}

\newcommand{\main}[1]{\textcolor{MainBlue}{#1}}
\newcommand{\lookahead}[1]{\textcolor{LookaheadRed}{#1}}

\newcommand{\modif}[1]{#1}

\newcommand{\fullwidthtrainingtable}[1]{%
  \begingroup
  \sbox0{#1}%
  \setlength{\tabcolsep}{\dimexpr\tabcolsep+\wd0/108\relax}%
  \resizebox{\textwidth}{!}{#1}%
  \endgroup
}

\newcommand{\fullwidthsmalltable}[1]{%
  \begingroup
  \sbox0{#1}%
  \setlength{\tabcolsep}{%
    \dimexpr\tabcolsep+(\textwidth-\wd0)/12\relax}%
  #1%
  \endgroup
}

\usepackage{xcolor}
\definecolor{SpringGreen}{RGB}{0,255,127}
\usepackage{soul} 

\definecolor{oursrow}{gray}{0.93}

\usepackage{pifont}
\usepackage{algorithm}
\usepackage{algpseudocode}
\usepackage{amsmath}
\usepackage{multicol}
\usepackage{setspace}
\usepackage{xcolor}
\definecolor{NiceGreen}{RGB}{34,139,34}   
\definecolor{NiceRed}{RGB}{220,50,47}    
\definecolor{stageA}{RGB}{255,236,209}  
\definecolor{stageB}{RGB}{215,236,251}  
\newcommand{\qmark}{\textcolor{black}{?}}
\newcommand{\cmark}{\textcolor{NiceGreen}{\ding{51}}}
\newcommand{\xmark}{\textcolor{NiceRed}{\ding{55}}}
\usepackage{tikz}
\usetikzlibrary{arrows.meta}
\definecolor{stageAA}{RGB}{230,145,30}  
\definecolor{stageBB}{RGB}{40,120,200} 
\definecolor{cfresh}{RGB}{120,119,112} 
\definecolor{cguide}{RGB}{136,135,128} 

\usepackage[most]{tcolorbox}

\usepackage[utf8]{inputenc} 
\usepackage[T1]{fontenc}    
\usepackage{hyperref}       
\usepackage{url}            
\usepackage{amsfonts}       
\usepackage{nicefrac}       
\usepackage{microtype}      
\usepackage{xcolor}         

\newcommand{\ours}{LACCO}

\usepackage{amsmath,amsthm,amsfonts}
\usepackage{algorithm}
\usepackage{algpseudocode}
\usepackage{amssymb}
\newtheorem{lemma}{Lemma}

\newtheorem{assumption}{Assumption}
\newtheorem{theorem}{Theorem}

\newtheorem{apptheorem}{Theorem}

\DeclareMathOperator{\InnerOpt}{InnerOpt}
\DeclareMathOperator{\OuterOpt}{OuterOpt}

\newcommand{\edouard}[1]{{\sethlcolor{SpringGreen}\hl{EO: #1}}}

\newcommand{\daniele}[1]{{\sethlcolor{yellow}\hl{D: #1}}}

\title{
\ours{}: Local Accumulation with Computation–Communication Overlap}

\author{
  Daniele Kopyshevskiy\textsuperscript{1} \quad
  Benjamin Th\'erien\textsuperscript{2,3} \quad
  Nicolas Pinto\textsuperscript{4,5} \quad
  Edouard Oyallon\textsuperscript{1,6}
  \\[0.8em]
  \makebox[\textwidth][c]{%
    \textsuperscript{1}Sorbonne Universit\'e, ISIR \quad
    \textsuperscript{2}MILA \quad
    \textsuperscript{3}Universit\'e de Montr\'eal
  }
  \\
  \makebox[\textwidth][c]{%
    \textsuperscript{4}Cygni Labs \quad
    \textsuperscript{5}Keytsugi \quad
    \textsuperscript{6}CNRS
  }
  \\[0.6em]
  \makebox[\textwidth][c]{%
    \texttt{daniele.kopyshevskiy@sorbonne-universite.fr}
  }
}

\if False
\author{%
  Daniele Kopyshevskiy\textsuperscript{1} \quad
  Benjamin Th\'erien\textsuperscript{2,3} \quad
  Nicolas Pinto\textsuperscript{4,5} \quad
  Edouard Oyallon\textsuperscript{1,6} \\[0.5em]
  \textsuperscript{1}Sorbonne Universit\'e, ISIR \quad
  \textsuperscript{2}MILA \quad
  \textsuperscript{3}Universit\'e de Montr\'eal \quad
  \textsuperscript{4}Cygni Labs  \quad
  \textsuperscript{5}Keytsugi \\
  \textsuperscript{6}CNRS 
  \texttt{benjamin.therien@umontreal.ca}, \texttt{edouard.oyallon@cnrs.fr}
}
\fi

\begin{document}

\definecolor{MainBlue}{RGB}{78,121,167}
\definecolor{LookaheadRed}{RGB}{196,100,93}

\maketitle

\begin{abstract}
    Training language models across low-bandwidth networks requires reducing synchronization overhead. Local optimization methods such as DiLoCo communicate only after several local steps, but outer gradient averaging and the outer update remain on the critical path. Existing computation--communication overlap methods avoid this stall through delayed or replica-local updates, which can degrade convergence or cause replicas to drift. We propose \ours{} (\textbf{L}ocal \textbf{A}ccumulation with \textbf{C}omputation–\textbf{C}ommunication \textbf{O}verlap), a two-stage algorithm that overlaps global communication with local computation while applying globally aggregated updates shared by every replica, without introducing staleness in the outer update. We establish convergence guarantees for smooth non-convex objectives. Experiments on language-model pre-training with FineWeb show that \ours{} consistently outperforms communication-overlapped baselines such as Eager and Delayed DiLoCo. These gains remain robust as model size scales up to \modif{0.5B} parameters and across a range of communication intervals and numbers of replicas up to 16 workers, while incurring only negligible—or fully overlappable—computational or memory overhead, which demonstrates the strength of \ours{}.
\end{abstract}

\section{Introduction}
Training large language models (LLMs) at scale requires distributing computation across hundreds or thousands of accelerators. This is typically achieved through multi-dimensional parallelism~\citep{shoeybi2019megatron,narayanan2021efficient}, which balances computation, communication, and memory requirements~\citep{narayanan2021efficient,rajbhandari2020zero}. For example, in distributed data-parallel training (DDP), multiple workers process different mini-batches concurrently and periodically synchronize their updates. At large scale, however, this synchronization can become a major bottleneck~\citep{li2020pytorch,rajbhandari2020zero}, as communicating gradients or parameters across workers is costly, particularly when training spans multiple clusters or geographically distributed data centers~\citep{douillard2023diloco,diskin2021distributed}. A broad range of techniques has been developed to reduce this communication overhead. Gradient compression methods, such as quantization~\citep{alistarh2017qsgd} and sparsification~\citep{aji2017sparse,ahmed2026know}, reduce the volume of data exchanged at each synchronization step. \modif{Beyond standard compression, zeroth-order optimization methods can reduce communication costs by transmitting scalar directional-derivative estimates and reconstructing perturbation directions from shared random seeds~\citep{qin2024federated, li2025achieving, feng2024baffle}, while LoRA-based methods communicate compact low-rank factors~\citep{hu2022lora, wang2024flora, babakniya2023slora}.} An alternative and complementary approach is to communicate less frequently: Local SGD~\citep{stich2019local,lin2020dont} follows this principle by allowing replicas to perform several local optimization steps before synchronizing their models. Besides reducing communication frequency, this approach can make distributed training more practical in heterogeneous or bandwidth-constrained environments~\citep{mcmahan2017communication,douillard2023diloco,cagnasso2026unifyinglocalcommunicationslocal}.

Yet, reducing the frequency of synchronization does not remove communication from the critical path. In conventional  training, workers still remain idle while waiting for a global communication round to complete~\citep{dean2012large,li2020pytorch}. 
Modern distributed training systems partially mitigate this  by overlapping communication with computation, transmitting each parameter or gradient shard~\citep{rajbhandari2020zero,douillard2025streaming} as soon as it becomes available~\citep{li2020pytorch,pytorch2020ddp}. 
By exploiting this natural staggering, these approaches reduce synchronization stalls without altering the underlying optimization.
Overlapping communication in Local SGD~\citep{stich2019local} raises a separate challenge: a communication round spans several local steps, during which local parameters keep evolving, so the communicated update may be stale by the time it is applied~\citep{liu2024asynchronous, kale2025eager}.
Streaming DiLoCo~\citep{douillard2025streaming}, for instance, staggers updates across parameter fragments, but merges each averaged update into local parameters that have already advanced, which in practice limits the overlap to a few  steps.
More generally, naively applying delayed updates degrades convergence~\citep{kale2025eager}, while compensating for staleness generally requires additional synchronization~\citep{kale2025eager}, memory~\citep{ren2021zerooffload}, or computation~\citep{zheng2017delay}.
In particular, methods designed primarily to overlap optimizer computation seek to mitigate the interaction between delayed global updates and multiple intervening local steps, rather than eliminating staleness altogether.

In this work, we pursue this latter direction by combining Local SGD~\citep{stich2019local} with a computation--communication overlap mechanism inspired by \citet{nabli2025acco}.
We introduce \ours{} (\textbf{L}ocal \textbf{A}ccumulation with \textbf{C}omputation–\textbf{C}ommunication \textbf{O}verlap), an algorithm that overlaps global communication with local optimization without introducing parameter staleness. Existing read--write locking approaches execute communication and computation on separate streams and therefore require two buffers to maintain distinct parameter states~\citep{nabli2025acco,ren2021zerooffload}. Standard methods~\citep{ren2021zerooffload} manage these buffers independently, effectively decoupling communication from local computation, yet introducing staleness. In contrast, the key idea behind \ours{} is to swap the communication and local-computation buffers, i.e. two parameter states, between two successive communication stages, while allowing each replica to make local progress. This design has the potential to hide communication latency behind computation while preserving a formulation that enables removal of staleness, and thereby good accuracy as well as simple convergence analysis.

 Our main contributions are as follows:  \textbf{(a) Replica consistency and no staleness.} We propose \ours{}, an algorithm that fully overlaps the outer step with local computation. Thanks to our novel two-stage mechanism for pseudo-gradients, replicas remain consistent, without staleness from outer updates. \textbf{(b) Theoretical convergence analysis.} We provide a theoretical convergence analysis for smooth non-convex objectives in the Local SGD setting. \textbf{(c) Consistent scaling.} Across models up to \modif{0.5B} parameters, communication intervals, replica counts, and matched compute budgets, \ours{} achieves the lowest evaluation loss among overlapping methods. It is the first method to fully overlap the outer step with local inner steps while remaining replica-consistent and staleness-free, improving significantly over DiLoCo variants designed for communication--computation overlap, while incurring a limited overhead. Our source code can be found at \href{https://github.com/kopyshevskiy/lacco}{https://github.com/kopyshevskiy/lacco}.

\section{Related Work}

\paragraph{Communication-Efficient Distributed Training.} 
The standard approach to training LLMs is DDP~\citep{li2020pytorch}, in which each worker maintains a full copy of the model, computes gradients independently, and synchronizes them at every iteration, leading to communication overhead. 
 One line of work mitigates this overhead by reducing the amount of information exchanged at each training step. 
Examples include compression~\citep{seide2014onebit, lin2018deep, janson2026stabilizing, team2026kimi}, quantization~\citep{alistarh2017qsgd, bernstein2018signsgd}, sparsification~\citep{aji2017sparse, wangni2018gradient}, \modif{low-rank updates~\citep{hu2022lora, wang2024flora, babakniya2023slora}, zeroth-order optimization using shared random seeds~\citep{qin2024federated, li2025achieving, feng2024baffle},} and decentralized synchronous~\citep{koloskova2020unified} and asynchronous~\citep{nabli2023dadao} methods that communicate only with a subset of neighboring replicas. These techniques can significantly reduce bandwidth requirements, but they modify the underlying optimization or communication procedure and may introduce additional variance. Our method is orthogonal to such approaches: it requires no modification to their underlying mechanisms and can be combined with them to further reduce communication overhead.

\paragraph{Local updates.}A complementary line of work, on which we build, reduces communication overhead by synchronizing less frequently~\citep{mcmahan2017communication}. In Local SGD~\citep{stich2019local}, replicas perform multiple local optimization steps before averaging their model parameters. FedOpt~\citep{reddi2021adaptive} generalizes this procedure through a bilevel optimization perspective: model averaging is interpreted as an outer server update, and the negative average displacement of the local models is treated as an outer gradient, or ``pseudo-gradient''. More recently, DiLoCo~\citep{douillard2023diloco,belilovsky2026inner} adapted this framework to LLM pre-training for low-bandwidth networks. It devotes significant design effort to the outer update, and in particular the choice of Nesterov SGD~\citep{nesterov1983method,sutskever2013importance} and its hyperparameters, yielding stronger empirical convergence than vanilla Local SGD. Nevertheless, while these methods reduce communication frequency, replicas must wait for the outer gradients to be aggregated and the outer update to be applied before beginning the next sequence of local steps. Our method extends  DiLoCo  by fully overlapping this outer iteration with the subsequent local computation, thereby hiding communication latency without introducing staleness.

\paragraph{Asynchronous Training.}
One way to hide communication latency is asynchronous training, which removes global synchronization entirely, and thus the straggler effect~\citep{dean2012large,lian2015asynchronous,nabli2023textbf}.
In parameter-server architectures~\citep{agarwal2011distributed,zhang2013asynchronous, mishchenko2022asynchronous}, each replica updates the global model on the server as soon as it completes a single step.
However, this introduces gradient staleness: gradients computed on an older version of the model are applied to a model that has since been updated by other workers. 
As a consequence, extensive research has addressed this via delay compensation~\citep{zheng2017delay}, polynomial staleness decay~\citep{xie2019asynchronous}, delay-dependent learning rates~\citep{barkai2019gap}, and weight-space lookahead correction~\citep{hakimi2019momentum}.
Recently, \citet{liu2024asynchronous} extended asynchrony to the DiLoCo framework through delayed Nesterov updates and dynamic local updates, showing that some existing asynchronous approaches~\citep{zheng2017delay,xie2019asynchronous} transfer poorly to this setting.
In a similar vein, \citet{ajanthan2025momentum} alleviate staleness via a lookahead delay correction mechanism that extrapolates in the negative direction of momentum.
All these methods apply delayed updates and then correct for them; in contrast, our two-stage mechanism never applies an outer gradient from a previous outer iteration, and thus avoids staleness by design.
Beyond staleness, Decoupled DiLoCo~\citep{decoupled2026diloco} uses asynchrony for fault tolerance, through quorum-based fragment aggregation that relaxes replica consistency to tolerate stragglers and failures.
Our method does not address stragglers or failures, which we leave for future work: \modif{merging our framework with asynchronous methods to handle network fluctuations and stragglers is entirely possible, with no fundamental restrictions in doing so, although we consider it beyond the scope of this work.}

\begin{table}[h]
\centering
\small
\caption{
\textbf{Comparison of outer computation--communication overlap methods among Local SGD methods.} \textbf{Full outer overlap} denotes overlapping a complete outer communication round with a local phase. \textbf{Replica consistency} means that replicas share identical outer model parameters after each outer update; \textbf{No ``local'' sync} means that the next outer update does not require outer gradients from the ongoing local phase.
}
\label{tab:overlap-comparison}

\begingroup
\renewcommand{\arraystretch}{1.18}
\setlength{\tabcolsep}{1.5pt}
\setlength{\extrarowheight}{2pt}

\begin{tabularx}{\textwidth}{
    @{}
    >{\raggedright\arraybackslash}p{0.21\textwidth}
    *{6}{>{\centering\arraybackslash}X}
    @{}
}
\toprule
\textbf{Method}
& \textbf{Comm quantity}
& \textbf{Full outer overlap}
& \textbf{Replica consistency}
& \textbf{No ``local'' sync}
& \textbf{Staleness free}
& \textbf{Convergence guarantees} \\
\midrule

Overlap-Local SGD
& Parameters & \xmark & \cmark & \xmark & \xmark & \cmark \\

CO2
& Parameters & \cmark & \cmark & \xmark & \xmark & \cmark \\

DiLoCo
& Outer grads & \xmark & \cmark & \cmark & \cmark & \cmark \\

Streaming DiLoCo
& Outer grads & \xmark & \xmark & \xmark & \xmark & \qmark \\

Delayed DiLoCo
& Outer grads & \cmark & \xmark & \cmark & \xmark & \qmark \\

Eager DiLoCo
& Outer grads & \cmark & \xmark & \xmark & \xmark & \qmark \\

\midrule
\rowcolor{oursrow}
\textbf{\ours{}}
& Outer grads & \cmark & \cmark & \cmark & \cmark & \cmark \\

\bottomrule
\end{tabularx}
\endgroup
\end{table}

\paragraph{Overlapping Communication and Computation.} We do not discuss the part of the literature that focuses on overlapping communication and computation across stages associated with other forms of parallelism, such as pipeline parallelism~\citep{liu2024deepseek,rivaud2025petra} or tensor parallelism~\citep{zhang2025ladder,singh2025model}. Instead, we restrict our focus to DDP. Unlike asynchronous methods, computation--communication overlap methods preserve an explicit outer-step structure: replicas perform local steps while communication is in flight and, once it completes, all receive the same update before proceeding.
Among them, an early framework is Overlap-Local SGD~\citep{wang2020overlap}, where replicas perform local steps and are then pulled toward an anchor model, computed as the average of the replicas delayed by one communication round.
Building on the same idea, CO2~\citep{sun2024co2} updates the global model with averaged parameters delayed by one outer iteration, but mitigates the resulting staleness through momentum clipping and a staleness-gap penalty.
Crucially, both methods introduce delays, which hurts convergence. 
To retain DiLoCo's outer momentum, Streaming DiLoCo~\citep{douillard2025streaming} overlaps outer gradient averaging of parameter subsets with subsequent local steps, and its authors report negligible degradation up to five steps. Similarly, \citet{maziane2026loscar}  relies on sparse model averaging with delay-corrected synchronization and is evaluated exclusively on non-LLM tasks. By contrast, our method widens the overlap to all local steps and does not introduce staleness. 
A simple way to achieve a full overlap with all the local steps, discussed by \citet{kale2025eager} as Delayed DiLoCo, is to use outer gradients delayed by one outer iteration. 
Nonetheless, applying only this delayed outer gradient comes at the cost of worse convergence.
To address this, Eager DiLoCo~\citep{kale2025eager} mixes each replica's fresh local outer gradient with the averaged outer gradient from the previous outer iteration.
While this improves training convergence, relying on each replica's own local outer gradient can introduce replica drift and requires a ``local'' synchronization step, as replicas must wait for the inner steps to finish before computing the outer iteration. 
Our method is compatible with this correction, but on its own it is more robust than Eager DiLoCo as the number of replicas increases and removes the need for this ``local'' synchronization step. 
Moreover, Factored Gossip DiLoCo~\citep{koneputugodage2026factored} similarly overlaps
communication, but by relaxing exact synchronization through a mixing/gossip step, a
setting we do not address.
Table~\ref{tab:overlap-comparison} summarizes the differences between the above methods. Finally, our work builds on \citet{nabli2025acco}, but differs in two important respects: ACCO can be viewed as operating in the $H=1$ regime, whereas our setting considers multiple local optimization steps. As a result, the standard ACCO formulation cannot directly rely on an inner--outer loop decomposition. Second, optimizer offloading is not feasible in our setting because the interval between consecutive gradient steps is too short to effectively hide the associated overhead. These differences require a substantially different treatment and constitute a significant extension beyond the original ACCO setting.

\section{Method}

\subsection{Local SGD Setup}

We consider the minimization of a differentiable objective $f : \mathbb{R}^d \to \mathbb{R}$ of the form
\begin{equation}
    \min_{\theta \in \mathbb{R}^d} f(\theta), \qquad f(\theta) = \mathbb{E}_{\xi \sim \mathcal{D}} \left[ F(\theta; \xi) \right],
\end{equation}
where $\theta \in \mathbb{R}^d$ denotes the global model, $\xi \sim \mathcal{D}$ is a random sample from a dataset $\mathcal{D}$, and $F(\theta; \xi)$ is the sample loss. We consider a distributed, data-homogeneous setting with $K$ replicas, each sampling data from the same distribution $\mathcal{D}$. Let $\theta_t \in \mathbb{R}^d$ denote the global model at outer iteration $t$, and $\theta_{t,h}^i \in \mathbb{R}^d$ denote the local model of replica $i$ after $h$ local optimization steps at outer iteration $t$. Initially, we assume that $\theta_{0,0}^i = \theta_0$ for all $i = 1, \dots, K$. Within each outer iteration $t$, replicas perform $H$ local steps using an inner optimizer, denoted by $\InnerOpt$, typically AdamW~\citep{loshchilov2019decoupled}. We denote these updates, for $\xi^i_{t,h}\sim \mathcal{D}$ independently sampled for each worker, 
\begin{equation}
    \theta_{t,h+1}^i = \InnerOpt\left(\theta_{t,h}^i, \nabla F\left(\theta_{t,h}^i; \xi_{t,h}^i\right)\right), \qquad h = 0, \dots, H-1,
\end{equation}
where the initialization $\theta_{t,0}^i$ for $t > 0$ depends on the specific method under consideration.

\subsection{Background}

\paragraph{DiLoCo.}
At each outer iteration $t$, replica $i$ initializes
$\smash{\theta_{t,0}^{i}}=\theta_t$ and performs $H$ inner steps.
It then computes its outer gradient $g_t^i$, which is averaged
across replicas to obtain $g_t$. The averaged gradient is applied
to $\theta_t$ using an outer optimizer, denoted by $\OuterOpt$. The resulting update is
\begin{equation}
    \begin{cases}
        g_t^i = \theta_t - \theta_{t,H}^i, \\
        \theta_{t+1} = \OuterOpt(\theta_t, g_t),
    \end{cases}
    \quad \text{with} \quad
    g_t = \frac{1}{K} \sum_{i=1}^{K} g_t^i.
\end{equation}
Synchronizing the outer gradient $g_t$ forces all replicas to be idle before
updating $\theta_{t+1}$ and starting the next $H$ inner steps. A simple way to
avoid this waiting is to use the outer gradient $g_{t-1}$ from the previous
outer iteration, which we discuss below.

\paragraph{Delayed DiLoCo.}
Delayed DiLoCo~\citep{kale2025eager} delays outer updates by one iteration
to overlap communication with computation. Each replica initializes
$\smash{\theta_{t,0}^i=\theta_t^i}$ (with equal
$\smash{\theta_0^i}$ across replicas), performs the inner steps, and
asynchronously communicates its local outer gradient. At $t=0$,
replicas proceed directly to the next inner steps, while for $t>0$, they
first apply the previously communicated gradient $g_{t-1}$ to
$\theta_{t,0}^i$. The resulting update is
\begin{equation}
    \begin{cases}
        g_t^i = \theta_t^i - \theta_{t,H}^i, \\
        \theta_{t+1}^i =
        \begin{cases}
            \theta_{t,H}^i, & t=0, \\
            \OuterOpt(\theta_t^i,g_{t-1}), & t>0,
        \end{cases}
    \end{cases}
    \quad \text{with} \quad
    g_{t-1} = \frac{1}{K}\sum_{i=1}^{K}g_{t-1}^i.
\end{equation}
This allows the communication of $g_t^i$ to overlap with the next local
steps, thereby hiding communication latency. However, applying the
one-step-delayed outer gradient $g_{t-1}$ to the current model introduces
staleness, which can degrade convergence in both local and non-local
training settings~\citep{kale2025eager,nabli2025acco,nabli2025decentralized}, as we will also see in Section~\ref{xps}.

\paragraph{Eager DiLoCo.}
To correct the staleness introduced by $g_{t-1}$, Eager DiLoCo~\citep{kale2025eager} replaces each replica's stale contribution with its fresh outer gradient $g_t^i$, available at the end of the $H$ inner steps on worker $i$. The resulting update is:
\begin{equation}
    \begin{cases}
        g_t^i = \theta_t^i-\theta_{t,H}^i, \\
        \theta_{t+1}^i =
        \begin{cases}
            \theta_{t,H}^i, & t=0, \\
            \OuterOpt(\theta_t^i,\hat g_t^i), & t>0,
        \end{cases}
    \end{cases}
    \quad \text{with} \quad
    g_{t-1} = \frac{1}{K}\sum_{i=1}^{K}g_{t-1}^i,
\end{equation}
where for $t > 0$, we define
\begin{equation}
    \hat g_t^i
    = g_{t-1}+\frac{1}{K}\left(g_t^i-g_{t-1}^i\right)=\frac 1K\big(g^i_t+\sum_{j\neq i} g^j_{t-1}).\label{eq:eager}
\end{equation}
Because the correction depends on the replica, replicas apply different outer updates and can drift apart over time. Our method, by contrast, preserves consistency across replicas and does not require the computation of local steps to be finished before applying the outer step.

\subsection{\ours: Local Accumulation with Computation–Communication Overlap}
\begin{figure}[h]
    \centering
    \includegraphics[
        width=\textwidth,
        height=0.20\textheight,
        keepaspectratio
    ]{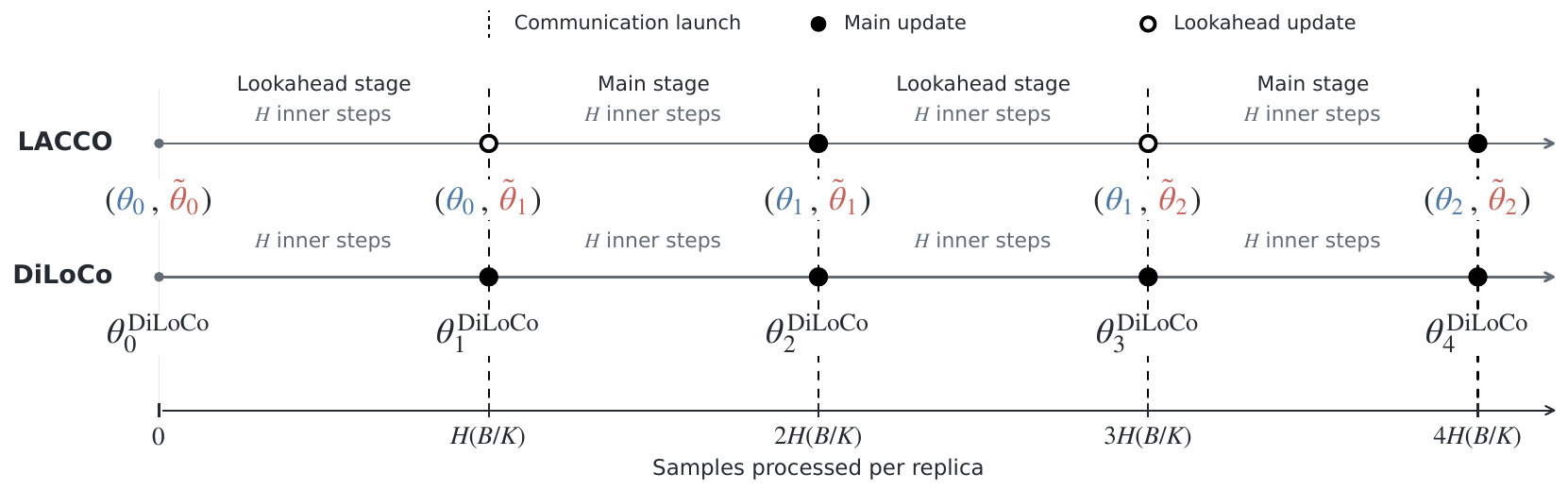}
    \caption{
        \textbf{Timeline of outer updates in \ours{} and DiLoCo.}
        Vertical marks indicate communication launches of outer gradients.
        $B$ is the global batch size and  $K$ is the number
        of replicas.
    }
    \label{fig:method-timeline}
    \vspace{-4mm}
\end{figure}

Following \citet{nabli2025acco}, and in contrast to DiLoCo variants, we maintain a pair of parameters $(\main{\theta_t},\lookahead{\tilde\theta_t})$, for both communication and computation. 
At each time step $t$, (\textit{i}) the roles of the two parameters  are swapped, after which (\textit{ii}) a lookahead stage and (\textit{iii}) a main stage are performed, each comprising $H$ inner steps. This yields the update sequence
$(\main{\theta_t},\lookahead{\tilde\theta_t}) \to
(\main{\theta_t},\lookahead{\tilde\theta_{t+1}}) \to
(\main{\theta_{t+1}},\lookahead{\tilde\theta_{t+1}})$.
As in DiLoCo~\citep{douillard2023diloco}, \ours{} communicates an outer gradient every $H$ inner steps, preserving the number of samples processed between communication rounds. We adjust the total number of steps accordingly so that the overall number of communications (and thus computations) remains unchanged. Figure~\ref{fig:method-timeline} highlights the resulting difference with DiLoCo. Accordingly, each replica maintains two inner optimizers, $\InnerOpt_{\mathrm{\main{M}}}$ and $\InnerOpt_{\mathrm{\lookahead{L}}}$, and two outer optimizers, $\OuterOpt_{\mathrm{\main{M}}}$ and $\OuterOpt_{\mathrm{\lookahead{L}}}$, associated with $\main{\theta_t}$ and $\lookahead{\tilde\theta_t}$, respectively.

\paragraph{Lookahead stage.}
Each replica $i$ initializes $\main{\theta_{t,0}^{i}}=\main{\theta_t}$ and performs $H$ inner steps, defined as
\begin{equation}
\main{\theta_{t,h+1}^{i}}
= \InnerOpt_{\mathrm{\main{M}}}\!\left(
\main{\theta_{t,h}^{i}},
\nabla F\!\left(\main{\theta_{t,h}^{i}};\xi_{t,h}^{i}\right)
\right),
\quad h=0,\ldots,H-1,
\end{equation}
where $\smash{\xi_{t,h}^{i}}\sim\mathcal{D}$.
It then computes its local outer gradient $\main{g_t^i}$ and communicates it asynchronously to obtain $\main{g_t}$. During these steps, communication of $\lookahead{\tilde g_t}$, computed from $\lookahead{\tilde \theta_t}$, completes. We apply this gradient to $\main{\theta_t}$ to construct the $\lookahead{\tilde \theta_{t+1}}$. 
Formally, the resulting update is thus
\begin{equation}
\begin{cases}
\main{g_t^i} = \main{\theta_t}-\main{\theta_{t,H}^{i}},\\
\lookahead{\tilde\theta_{t+1}} = \OuterOpt_{\mathrm{\lookahead{L}}}(\main{\theta_t},\lookahead{\tilde g_t}),
\end{cases}
\quad \text{with} \quad
\main{g_t}=\frac{1}{K}\sum_{i=1}^{K}\main{g_t^i}.
\end{equation}

\paragraph{Main stage.}
Each replica $i$ initializes $\lookahead{\tilde\theta_{t+1,0}^{i}}=\lookahead{\tilde\theta_{t+1}}$ and performs $H$ inner steps, defined as
\begin{equation}
\lookahead{\tilde\theta_{t+1,h+1}^{i}}
=
\InnerOpt_{\mathrm{\lookahead{L}}}\!\left(
\lookahead{\tilde\theta_{t+1,h}^{i}},
\nabla F\!\left(
\lookahead{\tilde\theta_{t+1,h}^{i}};
\tilde\xi_{t+1,h}^{i}
\right)
\right),
\qquad h=0,\ldots,H-1,
\end{equation}
where $\tilde\xi_{t+1,h}^{i}\sim\mathcal{D}$.
It then computes its local outer gradient $\lookahead{\tilde g_{t+1}^{i}}$ and launches its asynchronous communication. Once $\main{g_t}$ is available, we mix it with $\lookahead{\tilde g_t}$ to update the $\main{ \theta_{t+1}}$, so that
\begin{equation}
\begin{cases}
\lookahead{\tilde g_{t+1}^{i}}
= \lookahead{\tilde\theta_{t+1}}-\lookahead{\tilde\theta_{t+1,H}^{i}},\\
\main{\theta_{t+1}} = \OuterOpt_{\mathrm{\main{M}}}(\main{\theta_t},\frac 12(\main{g_t}+\lookahead{\tilde g_t})),
\end{cases}
\quad \text{with} \quad
\lookahead{\tilde g_{t+1}}
=\frac{1}{K}\sum_{i=1}^{K}\lookahead{\tilde g_{t+1}^{i}}.
\end{equation}

\begin{figure*}[h]
    \centering
    \includegraphics[width=\textwidth, height=0.30\textheight, keepaspectratio]{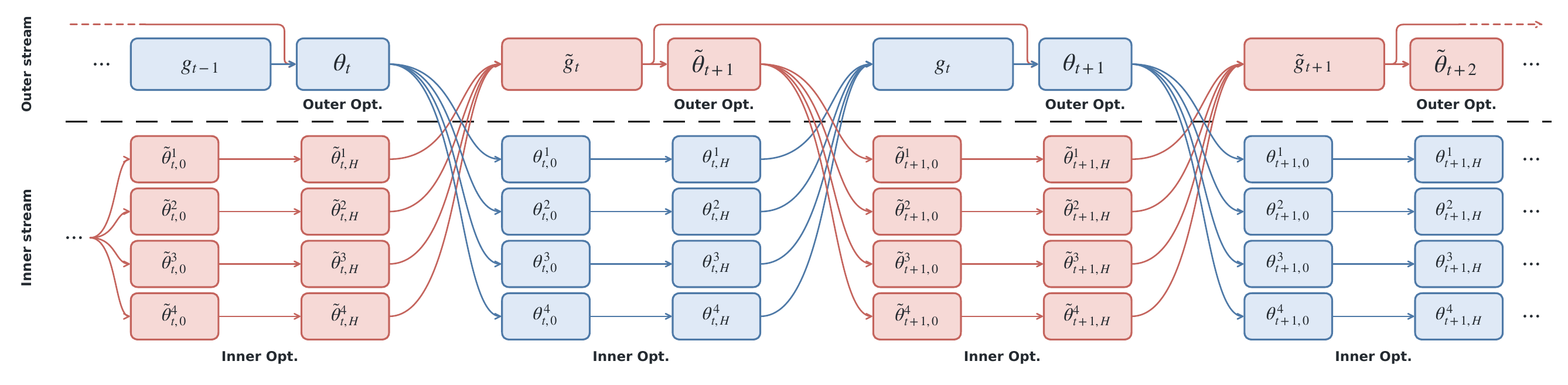}
    \caption{\textbf{Overview of \ours{}.}
    Each outer iteration alternates a lookahead stage (red) and a main stage (blue), each consisting of $H$ inner steps on $K=4$ replicas.
    While one stage performs its local steps, communication of the previous stage's outer gradient is overlapped with this computation.}
    \label{fig:method-overview}
    \vspace{-4mm}
\end{figure*}

\paragraph{Implementation.}
Each stage launches asynchronous communication of the outer gradient, overlapping it with local computation in the subsequent stage, as illustrated in Figure~\ref{fig:method-overview}. Each stage maintains separate outer and inner optimizer states; for the inner optimizers, we found it beneficial to average their momentum states at the end of each iteration~\modif{(see Appendix~\ref{sec:inner_optimizer_ablation})}. While maintaining two sets of parameters and optimizer states introduces additional memory overhead relative to DiLoCo and its variants, only one inner trajectory is active at a time. The inactive AdamW state can therefore be offloaded to CPU memory and prefetched before the next stage. The outer Nesterov momentum is similarly well suited for offloading, as it is accessed only during outer updates rather than at every local step. This allows most of the additional memory cost to be shifted from GPU to CPU memory, with only a small timing overhead, quantified in Section~\ref{mainres}. Complete pseudocode is provided in Appendix~\ref{app:algo}.

\subsection{Convergence Analysis}

In this section, for the sake of simplicity and without losing in generality, we establish the convergence of \ours{} when both
$\InnerOpt_{\mathrm{\main{M}}}$, $ \InnerOpt_{\mathrm{\lookahead{L}}}$
and $\OuterOpt_{\mathrm{\main{M}}}$, $ \OuterOpt_{\mathrm{\lookahead{L}}}$
are standard SGD. Let $K$ denote the number of replicas. For each $h=0,\ldots,H-1$ of iteration $t$, each replica $i$ draws samples $\smash{\xi^i_{t,h}}, \smash{\tilde\xi^i_{t,h}} \sim \mathcal{D}$ and performs the following local SGD
updates for $\beta>0$
\begin{equation}
\begin{aligned}
\theta_{t,0}^i &= \theta_t,
&\qquad
\theta_{t,h+1}^i
&= \theta_{t,h}^i
-\frac{\beta}{H}\nabla F(\theta_{t,h}^i;\xi_{t,h}^i), \\
\tilde\theta_{t,0}^i &= \tilde\theta_t,
&\qquad
\tilde\theta_{t,h+1}^i
&= \tilde\theta_{t,h}^i
-\frac{\beta}{H}\nabla F(\tilde\theta_{t,h}^i;\tilde\xi_{t,h}^i),
\end{aligned}
\qquad h=0,\dots,H-1.
\label{eq:inner}
\end{equation}
Let $\eta$ denote the outer learning rate. Then, the pair $\tilde\theta_t,\theta_t$ is updated as
\begin{equation}
\text{(i)}\quad
\begin{cases}
\tilde g_t^i = \tilde \theta_t-\tilde \theta_{t,H}^i, \\[4pt]
\displaystyle
\tilde g_t = \frac{1}{K}\sum_{i=1}^{K} \tilde g_t^i, \\[14pt]
\tilde\theta_{t+1} = \theta_t-\eta \tilde g_t,
\end{cases}
\qquad
\text{(ii)}\quad
\begin{cases}
 g_{t}^i
= \theta_{t}-\theta_{t,H}^i, \\[2pt]
\displaystyle
g_{t}
= \frac{1}{K}\sum_{i=1}^{K} g_{t}^i, \\[14pt]
\displaystyle
\theta_{t+1}
= \theta_t-\frac{\eta}{2}(g_t+\tilde g_{t}).
\end{cases}
\label{eq:sgd-dynamics}
\end{equation}
We make the following  assumptions.
\begin{assumption} \label{assumption:1} The objective $f:\mathbb{R}^d\to\mathbb{R}$ is bounded below by $f^\star$ and $L$-smooth, namely  $f^\star := \inf_{\theta\in\mathbb{R}^d} f(\theta) > -\infty$ and $\|\nabla f(\theta)-\nabla f(\theta')\| \le L\|\theta-\theta'\|$ for all $\theta,\theta'\in\mathbb{R}^d$.
\end{assumption}

\begin{assumption}
\label{assumption:2}
Let $\mathcal F_{t}$ denote the filtration generated by all
randomness preceding  iteration $t$.
Conditionally on $\mathcal F_t$, the local sample
sequences $\smash{(\xi_{t,0}^i,\ldots,\xi_{t,H-1}^i,\modif{\tilde\xi_{t,0}^i,\ldots,\tilde\xi_{t,H-1}^i})}$,
$i=1,\ldots,K$, are i.i.d. \modif{accross replicas.}
\end{assumption}

\begin{assumption}
\label{assumption:3}
Let $\mathcal F_{t,h}$ be defined as in
Assumption~\ref{assumption:2}.
The stochastic gradients satisfy
$\mathbb E[\nabla F(\theta_{t,h}^i;\xi_{t})\mid\mathcal F_{t,h}]
=\nabla f(\theta_{t,h}^i)$ and
$\mathbb E[\|\nabla F(\theta_{t,h}^i;\xi_{t})
-\nabla f(\theta_{t,h}^i)\|^2\mid\mathcal F_{t,h}]
\leq\sigma^2$.
\end{assumption}

Altogether, using standard Local SGD lemmas proved in Appendix \ref{app:proofs}, 
we can prove:

\begin{theorem}
\label{theorem:stochastic-convergence}
Suppose Assumptions~\ref{assumption:1}, \ref{assumption:2} and \ref{assumption:3} hold. Let \(H\geq1\) be the number of local SGD steps performed by $K$ replicas. Let \(\beta,\eta>0\) be as in~\eqref{eq:inner} and~\eqref{eq:sgd-dynamics}. Assume that
\modif{
\begin{equation}
0<\beta L\le\frac{1}{30},
\qquad
\eta=30.
\end{equation}
}
If \(\theta_0=\tilde\theta_0\), then, for every \(T\geq1\), we have
\modif{
\begin{equation}
\frac1T\sum_{t=0}^{T-1}
\mathbb E\left[
\|\nabla f(\theta_t)\|^2
\right]\le
\frac{
4(f(\theta_0)-f^\star)
}{15\beta T}+
\frac{824\beta L\sigma^2}{H}
+
\frac{18360\beta L\sigma^2}{KH}.
\label{eq:stochastic-convergence-bound}
\end{equation}
}
\end{theorem}
The key idea underlying the proof is that the following function serves
as a Lyapunov function for our dynamics adapted to Local SGD, by choosing,
\modif{
\begin{equation}
\mathcal V_t
=
f(\theta_t)-f^\star
+\eta \beta L\big(
f(\tilde\theta_t)-f^\star\big)
+
16L
\|\theta_t-\tilde\theta_t\|^2\geq0.
\end{equation}}
\section{Experiments}
\label{xps}

Our experiments use Llama $3$-style~\citep{dubey2024llama3} decoder-only
Transformers~\citep{vaswani2017attention} with $134$M\modif{, $325$M and $500$M} parameters. We use the Llama $2$ ($7$B) tokenizer~\citep{touvron2023llama2} with a vocabulary size of $32{,}000$ and a sequence length of $2{,}048$ tokens. Following standard practice~\citep{hoffmann2022training}, each model is trained on a token budget equal to $20\times$ its parameter count. We train on FineWeb~\citep{penedo2024fineweb} and validate on a disjoint held-out split of $65.5$M tokens. For methods where replicas are not synchronized, we average them before validation, while for \ours{} we validate on $\theta_t$~\modif{(see Appendix~\ref{sec:drift_ablation})}. Moreover, we perform zero-shot downstream tasks on classic benchmarks. Whereas~\cite{ajanthan2026asyncmesh} use $K=2$ replicas, we use $K\geq4$ replicas with a global batch size \modif{of $64$ for $134$M and $325$M models and $128$ for the $500$M model}. Architectural details, hyperparameter choices, and implementation details, all built on TorchFT~\citep{torchft}, are provided in Appendices~\ref{app:arch}, \ref{app:hyp}, and~\ref{app:impl}, respectively. We include DDP and DiLoCo as synchronous references and overlapping baselines use the same amount of computation and communication. 

\subsection{Main Results}
\label{mainres}

\paragraph{{\ours} evaluation on 134M model size with $K=4$ and $H=30$.} Table~\ref{tab:training_results_134m} compares \ours{} with both synchronous
baselines and methods that overlap communication with local computation.
Thanks to its two-stage mechanism, \ours{} achieves the lowest evaluation loss among communication-overlapping methods, and furthermore, \ours{} remains within $1.77\%$ of DDP and $0.91\%$ of DiLoCo. It also outperforms the overlapping baselines on HellaSwag and PIQA, and is close on ARC-Easy. \modif{While CPU-GPU transfers can be beneficial, our primary target setting is a tightly interconnected GPU cluster, where high-bandwidth communication enables the system to operate effectively as a single, unified computing platform. We however discuss below timing in a setting where this might not be the case.}

\begin{table}[h]
\centering
\small
\caption{
\textbf{134M evaluation, $K=4$ and $H=30$.}
Comparison of potentially overlapping baselines.
}
\label{tab:training_results_134m}

\fullwidthtrainingtable{%
\begin{tabular}{lccccc}
\toprule
\textbf{Method}
& \textbf{Overlap}
& \textbf{Eval loss} $\downarrow$
& \textbf{HellaSwag} $\uparrow$
& \textbf{PIQA} $\uparrow$
& \textbf{ARC-Easy} $\uparrow$ \\
\midrule

DDP
& \xmark & 2.936 & 28.76 & 59.79 & 38.80 \\

DiLoCo
& \xmark & 2.961 & 29.06 & 60.50 & 36.91 \\

\noalign{\vskip 2pt}
\hdashline
\noalign{\vskip 2pt}
\rowcolor{oursrow}
\ours{}
& \cmark & \textbf{2.988} & \textbf{28.87} & \textbf{61.10} & 37.04 \\

Eager DiLoCo
& \cmark & 3.020 & 28.77 & 59.74 & \textbf{37.50} \\

Delayed DiLoCo
& \cmark & 3.117 & 27.81 & 58.98 & 35.98 \\

\bottomrule
\end{tabular}%
}
\end{table}

\paragraph{Simulated wall-clock time.}
\begin{figure}[h]
\centering

\begin{subfigure}[h]{0.32\linewidth}
    \centering
    \includegraphics[width=\linewidth]{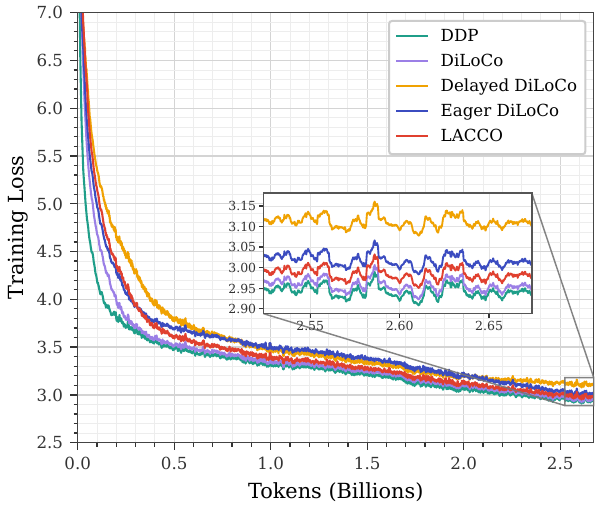}
    \caption{Measured loss versus tokens.}
    \label{fig:main-results-tokens}
\end{subfigure}
\hfill
\begin{subfigure}[h]{0.32\linewidth}
    \centering
    \includegraphics[width=\linewidth]
    {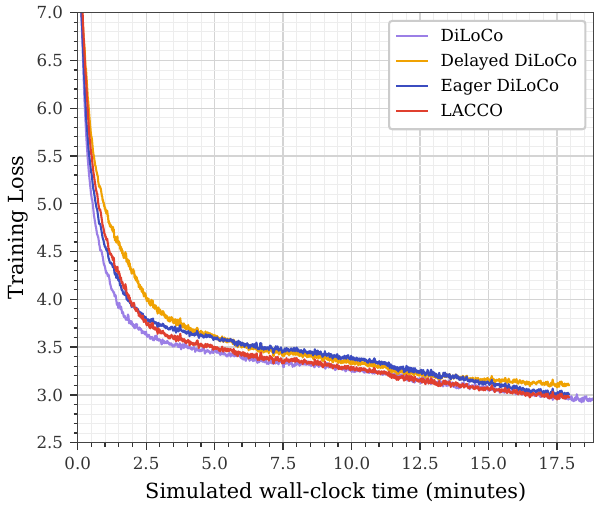}
    \caption{$\OuterOpt$ state on GPU.}
    \label{fig:main-results-gpu}
\end{subfigure}
\hfill
\begin{subfigure}[h]{0.32\linewidth}
    \centering
    \includegraphics[width=\linewidth]
    {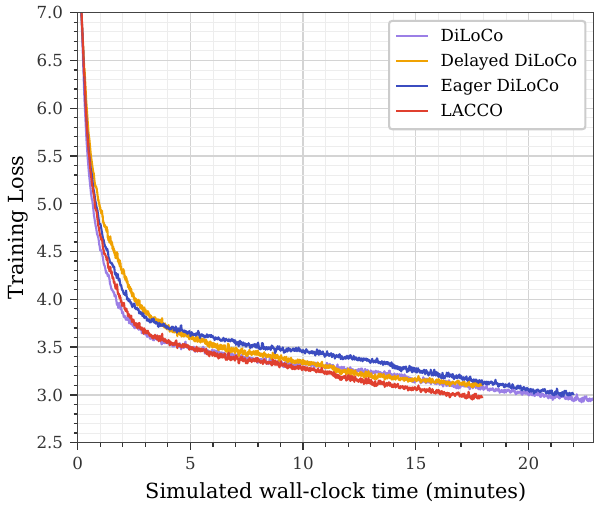}
    \caption{$\OuterOpt$ state offloaded.}
    \label{fig:main-results-cpu}
\end{subfigure}

\caption{
\textbf{Training loss versus tokens and simulated wall-clock time.}
(a) Measured loss versus tokens. (b)--(c) The same losses against simulated
time (H100, $50\%$ MFU, $b_{\mathrm{net}}=10\,\mathrm{GB/s}$,
$b_{\mathrm{CPU/GPU}}=3\,\mathrm{GB/s}$), with the
$\OuterOpt$ state kept on GPU (b) or offloaded to CPU (c).
}
\vspace{-2mm}
\label{fig:main-results}

\end{figure}

Figure~\ref{fig:main-results} reports training loss against tokens and
simulated wall-clock time using the timing model detailed in
Appendix~\ref{sec:timing_bandwidth}. We assume a pessimistic 
inter-replica bandwidth of $b_{\mathrm{net}}=10\,\mathrm{GB/s}$ and CPU--GPU bandwidth of
$b_{\mathrm{CPU/GPU}}=3\,\mathrm{GB/s}$. Let $T_{\mathrm{inner}}$, $T_{\mathrm{comm}}$, and $T_{\mathrm{outer}}$
denote the time spent on local computation, outer-gradient communication,
and the outer update, respectively. Without overlap, the time per round is
$T_{\mathrm{serial}}
=
T_{\mathrm{inner}}+T_{\mathrm{comm}}+T_{\mathrm{outer}}$. If communication is overlapped with local computation but the outer update
remains serialized,
\[
T_{\mathrm{partial}}
=
\max\{T_{\mathrm{inner}},T_{\mathrm{comm}}\}
+T_{\mathrm{outer}}.
\]
If the complete outer path is overlapped with local computation,
\[
T_{\mathrm{overlap}}
=
\max\{T_{\mathrm{inner}},
      T_{\mathrm{comm}}+T_{\mathrm{outer}}\}.
\]
Hence, communication and outer-update costs are fully hidden whenever
$T_{\mathrm{inner}}
\geq
T_{\mathrm{comm}}+T_{\mathrm{outer}}.$ For the $134$M model with $K=4$, $H=30$, and $131{,}072$ tokens per
inner step, we consider an H100 (details in Appendix~\ref{sec:timing_bandwidth}). This gives 
$T_{\mathrm{inner}}\simeq1.58\,\mathrm{s}$, and
$T_{\mathrm{comm}}\simeq0.08\,\mathrm{s}$. For \ours{}, the additional
optimizer-state transfers amount to an estimated
$T_{\mathrm{outer}}\simeq0.89\,\mathrm{s}$, whereas it is $T_{\mathrm{outer}}\simeq0.36\,\mathrm{s}$ for the other methods. The complete outer path is
therefore hidden by the local computation at $H=30$: under these
assumptions, full overlap requires only $H\geq19$, as detailed in Appendix~\ref{sec:timing_bandwidth}. This threshold is
independent of the parameter count $P$ under our simplified model:
both the local-computation cost and the communicated/offloaded volumes
scale linearly with $P$. This cancellation assumes fixed MFU and
effective bandwidths and neglects the comparatively small optimizer
kernel time. All model computation is assumed to execute on the GPU. 

\paragraph{\modif{Real wall-clock time.}}
\modif{To validate our timing model under more realistic scenarios, Table~\ref{tab:wallclock_real} reports empirical wall-clock measurements for $K=4$ workers training the 0.5B model at $H=30$ on 80GB NVIDIA A100 GPUs (PCIe Gen4 x16) across low inter-node bandwidths of 0.5\,Gbps, 1\,Gbps, and 5\,Gbps. We compare synchronous DiLoCo against two LACCO variants: (i) CPU-offloaded inactive optimizer states with GPU outer updates where $T_{\text{outer}} \simeq 0.6\,\text{s}$, and (ii) fully GPU-resident states where $T_{\text{outer}} \simeq 0.15\,\text{s}$, significantly shorter than the gradient steps or communications. 
Each inner step takes $T_{\mathrm{step}}\simeq1.72$\,s, giving $T_{\mathrm{inner}}=H T_{\mathrm{step}}\simeq51.6$\,s. 
Our timing model (Appendix~\ref{sec:timing_bandwidth}) gives $T_{\mathrm{round}}\simeq T_{\mathrm{inner}}+T_{\mathrm{comm}}+T_{\mathrm{outer}}$ for DiLoCo and $T_{\mathrm{round}}\simeq\max\{T_{\mathrm{inner}},T_{\mathrm{comm}}+T_{\mathrm{outer}}\}$ for LACCO.
While DiLoCo is severely bottlenecked by serialized synchronization, both LACCO implementations successfully hide the communication and outer update behind the total local compute window of $51.6\,\text{s}$. This yields an end-to-end training speedup of up to 1.89$\times$ at 0.5\,Gbps and 1.44$\times$ at 1\,Gbps, with identical wall-clock performance between the offloaded and GPU-resident configurations in this setting. Moreover, for a detailed breakdown of memory footprints and training throughput, please refer to Appendix~\ref{sec:memory_throughput_ablation}.}

\begin{table}[t]
\centering
\small
\caption{\modif{
Measured wall-clock breakdown and end-to-end speedup with $K=4$ workers and $H=30$ on the 0.5B model. $T_{\mathrm{step}}$ denotes the time per inner step, $T_{\mathrm{inner}}=H T_{\mathrm{step}}$ the total local-computation time, and $T_{\mathrm{round}}$ the measured time per communication round.
}}
\label{tab:wallclock_real}
\resizebox{\linewidth}{!}{%
\begin{tabular}{llcccccc}
\toprule
\modif{Bandwidth} & \modif{Method} & \modif{$T_{\mathrm{step}}$ (s)} & \modif{$T_{\mathrm{inner}}$ (s)} & \modif{$T_{\mathrm{comm}}$ (s)} & \modif{$T_{\mathrm{outer}}$ (s)} & \modif{$T_{\mathrm{round}}$ (s)} & \modif{Speedup} \\
\midrule
\multirow{3}{*}{\modif{0.5 Gbps}}
 & \modif{DiLoCo}             & \modif{1.72} & \modif{51.6} & \modif{49.4} & \modif{0.3}  & \modif{101.17} & \modif{1.00$\times$} \\
 & \modif{LACCO, CPU offload} & \modif{1.72} & \modif{51.6} & \modif{49.4} & \modif{0.6}  & \modif{54.05}  & \modif{1.87$\times$} \\
 & \modif{LACCO, fully GPU}   & \modif{1.72} & \modif{51.6} & \modif{49.4} & \modif{0.15} & \modif{53.62}  & \modif{1.89$\times$} \\
\midrule
\multirow{3}{*}{\modif{1 Gbps}}
 & \modif{DiLoCo}             & \modif{1.72} & \modif{51.6} & \modif{24.8} & \modif{0.3}  & \modif{76.38}  & \modif{1.00$\times$} \\
 & \modif{LACCO, CPU offload} & \modif{1.72} & \modif{51.6} & \modif{24.8} & \modif{0.6}  & \modif{53.21}  & \modif{1.44$\times$} \\
 & \modif{LACCO, fully GPU}   & \modif{1.72} & \modif{51.6} & \modif{24.8} & \modif{0.15} & \modif{53.08}  & \modif{1.44$\times$} \\
\midrule
\multirow{3}{*}{\modif{5 Gbps}}
 & \modif{DiLoCo}             & \modif{1.72} & \modif{51.6} & \modif{5.3}  & \modif{0.3}  & \modif{56.66}  & \modif{1.00$\times$} \\
 & \modif{LACCO, CPU offload} & \modif{1.72} & \modif{51.6} & \modif{5.3}  & \modif{0.6}  & \modif{53.00}  & \modif{1.07$\times$} \\
 & \modif{LACCO, fully GPU}   & \modif{1.72} & \modif{51.6} & \modif{5.3}  & \modif{0.15} & \modif{52.95}  & \modif{1.07$\times$} \\
\bottomrule
\end{tabular}%
}
\end{table}

\subsection{Ablations}

\subsubsection{Varying the model size}

\begin{table}[h]
\centering
\small
\caption{
\textbf{Model size evaluation, $K=4$ and $H=30$.}
Comparison across $134$M, \modif{$325$M and $500$M} models. \ours~consistently outperforms other overlapping methods and is on par with  DiLoCo.
}
\label{tab:training_results}

\fullwidthtrainingtable{%
\begin{tabular}{llccccc}
\toprule
\textbf{Scale}
& \textbf{Method}
& \begin{tabular}[c]{@{}c@{}}
    \textbf{Overlap}
  \end{tabular}
& \textbf{Eval loss} $\downarrow$
& \textbf{HellaSwag} $\uparrow$
& \textbf{PIQA} $\uparrow$
& \textbf{ARC-Easy} $\uparrow$ \\
\midrule

134M
& DDP            & \xmark & 2.936 & 28.76 & 59.79 & 38.80 \\
& DiLoCo         & \xmark & 2.961 & 29.06 & 60.50 & 36.91 \\
\noalign{\vskip 2pt}
\hdashline
\noalign{\vskip 2pt}
\rowcolor{oursrow}
& \ours{}        & \cmark & \textbf{2.988} & \textbf{28.87} & \textbf{61.10} & 37.04 \\
& Eager DiLoCo   & \cmark & 3.020 & 28.77 & 59.74 & \textbf{37.50} \\
& Delayed DiLoCo & \cmark & 3.117 & 27.81 & 58.98 & 35.98 \\

\midrule

325M
& DDP            & \xmark & 2.673 & 33.90 & 64.80 & 42.76 \\
& DiLoCo         & \xmark & 2.692 & 34.27 & 64.25 & 41.20 \\
\noalign{\vskip 2pt}
\hdashline
\noalign{\vskip 2pt}
\rowcolor{oursrow}
& \modif{\ours{}}        & \cmark & \modif{\textbf{2.693}} & \modif{\textbf{34.12}} & \modif{63.49} & \modif{41.37} \\
& \modif{Eager DiLoCo}   & \cmark & \modif{2.715} & \modif{33.78} & \modif{\textbf{64.09}} & \modif{\textbf{42.26}} \\
& Delayed DiLoCo & \cmark & 2.795 & 31.39 & 63.11 & 39.44 \\

\midrule

\modif{500M} & \modif{DDP}            & \xmark & \modif{2.558} & \modif{37.31} & \modif{66.16} & \modif{45.50} \\
     & \modif{DiLoCo}         & \xmark & \modif{2.557} & \modif{38.53} & \modif{66.05} & \modif{46.34} \\
\noalign{\vskip 2pt}
\hdashline
\noalign{\vskip 2pt}
\rowcolor{oursrow}
     & \modif{\ours{}}          & \cmark & \modif{\textbf{2.561}} & \modif{\textbf{38.63}} & \modif{66.10} & \modif{45.58} \\
     & \modif{Eager DiLoCo}   & \cmark & \modif{2.576} & \modif{37.88} & \modif{\textbf{66.65}} & \modif{\textbf{45.62}} \\
     & \modif{Delayed DiLoCo} & \cmark & \modif{2.675} & \modif{34.12} & \modif{64.53} & \modif{42.97} \\

\bottomrule
\end{tabular}%
}
\end{table}

We train and evaluate all methods at \modif{$134$M, $325$M and $500$M} parameters using $H=30$ and $K=4$, and we tune the baselines according to Appendix \ref{app:hyp}. In Table~\ref{tab:training_results}, we observe that \ours{} achieves the lowest evaluation loss among communication-overlapping methods at every model size, demonstrating consistent scaling. At the larger \modif{$500$M} scale, \ours{} achieves a validation
loss only approximately \modif{$0.11\%$} higher than DDP while communicating (i)
$30\times$ less frequently and (ii) fully hiding communication latency, with (iii) comparable downstream accuracy.



\subsubsection{Varying the inner steps $H$}

\begin{table}[ht]
\centering
\small
\caption{
\textbf{Evaluation of the inner step size $H$, with model size 134M and $K = 4$.} DiLoCo, \ours{}, and Eager DiLoCo all show a U-shaped trend in $H$, as observed in \citep{kale2025eager}. Only Delayed DiLoCo does not exhibit this trend, which is consistent with its delayed update dynamics.
}
\label{tab:scaling_h}
\fullwidthtrainingtable{%
\begin{tabular}{lccccccc}
\toprule
\textbf{Method}
& $H=5$ & $H=10$ & $H=15$ & $H=30$ & $H=45$ & $H=60$ & $H=75$ \\
\midrule
DiLoCo         & 3.034 & 2.959 & 2.960 & 2.961 & 2.984 & 3.000 & 3.016 \\
\hdashline
\noalign{\vskip 2pt}
\rowcolor{oursrow}
\ours{}        & 3.077 & \textbf{2.960} & \textbf{2.959} & \textbf{3.001} & \textbf{3.039} & \textbf{3.071} & \textbf{3.103} \\
Eager DiLoCo   & 4.370 & 3.170 & 3.068 & 3.020 & 3.053 & 3.079 & 3.114 \\
Delayed DiLoCo & \textbf{3.040} & 3.061 & 3.074 & 3.117 & 3.150 & 3.176 & 3.201 \\
\bottomrule
\end{tabular}%
}
\end{table}

We vary $H$ from 5 to 75 on the 134M model with $K=4$ (Table~\ref{tab:scaling_h}) and keep the other hyperparameters constant. \ours{} achieves the lowest loss among overlapping methods for every $H > 5$, and performs comparably to synchronous DiLoCo for $H=10$ and $H=15$. At larger $H$, the gap to DiLoCo widens, possibly because $\theta_t$ and $\tilde\theta_t$ drift further apart over longer local phases~\modif{(see Appendix~\ref{sec:drift_ablation})}.

\subsubsection{Varying the number of replicas $K$}

\begin{table}[h]
\centering
\small
\caption{
\textbf{Evaluation of $K$, for 134M scale and $H = 30$.}
 Loss
degrades with $K$ for all methods,
with $K>8$ known to be a critical regime for DiLoCo~\citep{douillard2023diloco}. Nonetheless,
\ours{} remains within $1\%$ of DiLoCo at every $K$, degrading more
gracefully than other overlapped methods.
}
\label{tab:scaling_k}
\fullwidthsmalltable{%
\begin{tabular}{lcccc}
\toprule
\textbf{Method}
& $K=2$ & $K=4$ & $K=8$ & $K=16$ \\
\midrule
DiLoCo         & 2.941 & 2.961 & 3.029 & 3.143 \\
\hdashline
\noalign{\vskip 2pt}
\rowcolor{oursrow}
\ours{}        & 2.953 & \textbf{2.988} & \textbf{3.054} & \textbf{3.154} \\
Eager DiLoCo   & \textbf{2.946} & 3.020 & 3.183 & 3.445 \\
Delayed DiLoCo & 3.027 & 3.117 & 3.209 & 3.353 \\
\bottomrule
\end{tabular}%
}
\end{table}

We vary $K$ from 2 to 16 on the 134M model (by keeping the global batch size fixed) with $H=30$ (Table~\ref{tab:scaling_k}) and keep the other hyperparameters constant. \ours{} remains within $1\%$ of DiLoCo for all $K$ and achieves the lowest loss among overlapping methods for $K\geq4$.
In contrast, Eager DiLoCo degrades markedly as $K$ grows, from $3.020$ at $K=4$ to $3.445$ at $K=16$, compared with $2.988$ to $3.154$ for \ours{}. 
Eager DiLoCo is slightly better at $K=2$: with a single replica, local and averaged outer gradients coincide, and with two replicas each local outer gradient still accounts for half of the average. As $K$ increases, the discrepancy between replica-local and global updates grows, whereas the shared updates of \ours{} keep all replicas consistent. Consistently with Equation \ref{eq:eager}, the gap between Eager and Delayed DiLoCo vanishes as $K$ grows. 

\section{Conclusion}

We introduced \ours{}, a two-stage local optimization algorithm that fully overlaps outer communication and the outer update with local computation. Unlike existing overlap methods, \ours{} applies the same globally averaged update to every replica and computes each outer update from the current pair of models, thereby preserving replica consistency without introducing staleness or local synchronization. We established convergence guarantees for smooth  objectives and showed that, among communication-overlapping methods, \ours{} achieves the lowest evaluation loss across model sizes, communication intervals, and replica counts, while remaining close to synchronous DiLoCo and scaling markedly better than Eager DiLoCo as the number of replicas grows.

\section{AI disclosure}
In this work, we used generative AI tools to polish the manuscript and
assist with code review and debugging. All AI-assisted edits and code
changes were reviewed by the authors. We take responsibility for the
final content of this work, including all text, claims, and code produced
with the assistance of generative AI.

\ificlrfinal
\subsubsection*{Acknowledgments}
B.T. acknowledges support from the FRQNT Doctoral (B2X) scholarship. D.K. and E.O. were supported by funding managed by the French National Research Agency (ANR) under the France 2030 program through the  PEPR IA (grant
SHARP ANR-23-PEIA-0008). Part of this work was
granted access to the JeanZay HPC/AI resources of IDRIS under the allocation AD011015884R1, AD011017481, AD011016641, AD011017661 and AD011017766.
\fi

\bibliography{iclr2027_conference}
\bibliographystyle{iclr2027_conference}


\newpage

\appendix

\section{Algorithmic View}

\begin{algorithm}[ht]
\caption{\ours}
\label{app:algo}
\begin{algorithmic}[1]

\Statex \textbf{Inputs:}
number of outer iterations $T$,
inner steps $H$, number of replicas $K$
\Statex \textbf{Inputs:}
initial models
$\main{\theta_0}=\lookahead{\tilde{\theta}_0}$,
initial $\lookahead{\tilde{g}_0} = 0$
\Statex \textbf{Inputs:} optimizers
$\InnerOpt_{\mathrm{\main{M}}}$, $\InnerOpt_{\mathrm{\lookahead{L}}}$,
$\OuterOpt_{\mathrm{\main{M}}}$, $\OuterOpt_{\mathrm{\lookahead{L}}}$
\Statex \textbf{Inputs:} data shards $\mathcal{D}_1,\ldots,\mathcal{D}_K$

\Function{\textsc{InnerSteps}}{$\vartheta_0^i,\mathcal{D}_i,H,\mathrm{Opt}$}
    \For{$h=0,\ldots,H-1$}
        \State $\xi_h^i \sim \mathcal{D}_i$
        \State $\vartheta_{h+1}^i
        \gets
        \mathrm{Opt}\!\left(
            \vartheta_h^i,
            \nabla F(\vartheta_h^i;\xi_h^i)
        \right)$
    \EndFor
    \State \Return $\vartheta_H^i$
\EndFunction

\For{$t=0,\ldots,T-1$}
    \ForAll{$i\in\{1,\ldots,K\}$ \textbf{in parallel}}

        \Statex \hspace{\algorithmicindent}\textit{Lookahead stage}
        \State $\main{\theta_{t,H}^i}
        \gets
        \Call{\textsc{InnerSteps}}{
            \main{\theta_t},\mathcal{D}_i,H,\InnerOpt_{\mathrm{\main{M}}}
        }$
        \State $\main{g_t^i}
        \gets
        \main{\theta_t}-\main{\theta_{t,H}^i}$
        \State $\main{g_t}
        \gets
        \textsc{Async-AllReduce}(\main{g_t^i})$
        \State $\textsc{Wait}(\lookahead{\tilde{g}_t})$
        \State $\lookahead{\tilde{\theta}_{t+1}}
        \gets
        \OuterOpt_{\mathrm{\lookahead{L}}}\!\left(
            \main{\theta_t},
            \lookahead{\tilde{g}_t}
        \right)$

        \Statex \hspace{\algorithmicindent}\textit{Main stage}
        \State $\lookahead{\tilde{\theta}_{t+1,H}^i}
        \gets
        \Call{\textsc{InnerSteps}}{
            \lookahead{\tilde{\theta}_{t+1}},
            \mathcal{D}_i,
            H,
            \InnerOpt_{\mathrm{\lookahead{L}}}
        }$
        \State $\lookahead{\tilde{g}_{t+1}^i}
        \gets
        \lookahead{\tilde{\theta}_{t+1}}
        -
        \lookahead{\tilde{\theta}_{t+1,H}^i}$
        \State $\lookahead{\tilde{g}_{t+1}}
        \gets
        \textsc{Async-AllReduce}(
            \lookahead{\tilde{g}_{t+1}^i}
        )$
        \State $\textsc{Wait}(\main{g_t})$
        \State $\main{\theta_{t+1}}
        \gets
        \OuterOpt_{\mathrm{\main{M}}}\!\left(
            \main{\theta_t},
            \frac 12(\,\main{g_t}
            +
            \,\lookahead{\tilde{g}_t})
        \right)$

    \EndFor
\EndFor

\end{algorithmic}
\end{algorithm}

\section{Experimental Details}

\label{app:modelarchandhyp}

\subsection{Model architecture}\label{app:arch}

Our experiments use Llama 3-style~\citep{dubey2024llama3} decoder-only Transformers
\citep{vaswani2017attention} with RMSNorm~\citep{zhang2019rmsnorm} pre-normalization,
SwiGLU feed-forward networks~\citep{shazeer2020glu}, and rotary positional embeddings
(RoPE)~\citep{su2021roformer} with base $\theta = 10{,}000$. 
Attention uses standard
multi-head attention with head dimension $64$. 
We use the Llama 2 ($7$B) tokenizer
\citep{touvron2023llama2} with a vocabulary size of $32{,}000$ and a sequence length
of $2{,}048$ tokens. Following the Chinchilla scaling laws~\citep{hoffmann2022training},
each model is trained on a token budget equal to $20\times$ its parameter count. All
models are trained from scratch, with weights initialized following
\citet{dubey2024llama3}. 
Details of the
architectures used at each scale are reported in Table~\ref{tab:architectures}.

\begin{table}[H]
\centering
\small
\caption{
\textbf{Model architecture.} We report, for each scale, the total parameter
count, the number of Transformer layers, the number of attention heads, the
embedding dimension, the per-head dimension, the SwiGLU feed-forward hidden
dimension, and the Chinchilla token budget.}
\label{tab:architectures}

\renewcommand{\arraystretch}{1.15}
\setlength{\tabcolsep}{8pt}

\begin{tabular}{@{}lcccccc@{}}
\toprule
\textbf{\shortstack[l]{Model\\parameters}}
& \shortstack{Transformer\\layers}
& \shortstack{Attention\\heads}
& \shortstack{Embedding\\dimension}
& \shortstack{Head\\dimension}
& \shortstack{FFN hidden\\dimension}
& \shortstack{Token\\budget}
\\
\midrule
134M & 12 & 12 & 768  & 64 & 2048 & 2.7B  \\
325M & 20 & 16 & 1024 & 64 & 2816 & 6.5B  \\
\modif{500M} & \modif{20} & \modif{20} & \modif{1280} & \modif{64} & \modif{3584} & \modif{9.8B}  \\
\bottomrule
\end{tabular}
\end{table}

\subsection{Hyperparameter Tuning}\label{app:hyp}
We use AdamW~\citep{loshchilov2019decoupled} as the inner optimizer with
$\beta_1=0.9$ and $\beta_2=0.95$. 
The inner learning rate follows a cosine
decay schedule~\citep{loshchilov2017sgdr} with a $10\%$ linear warm-up
\citep{goyal2017accurate} \modif{from $1\%$ of its peak value}. 
Weight decay is set to $0.1$. 
The outer optimizer is Nesterov SGD~\citep{nesterov1983method,sutskever2013importance}.
\modif{All hyperparameters were tuned on the $134$M model with $H=30$ and $K=4$. When scaling to the $325$M and $500$M models, the outer learning rate and momentum are kept fixed for each method~\citep{charles2025scaling}.} 
\modif{We use a global batch size of $B=64$ for the $134$M and $325$M models and $B=128$ for the $500$M model.}
For DDP, the inner learning rate $\eta_{\mathrm{in}}$ was swept over
$ \eta_{\mathrm{in}} \in \{1\mathrm{e}{-3}, 2\mathrm{e}{-3}, 3\mathrm{e}{-3}, 4\mathrm{e}{-3},
5\mathrm{e}{-3}, 6\mathrm{e}{-3}\}$.
For all the other methods, the inner learning rate $\eta_{\mathrm{in}}$ was swept over
$\{6\mathrm{e}{-4}, 7\mathrm{e}{-4}, 8\mathrm{e}{-4}, 1\mathrm{e}{-3},
2\mathrm{e}{-3}, 3\mathrm{e}{-3}\}$, the outer learning rate $\eta_{\mathrm{out}}$ over
$\{0.2, 0.3, 0.4, 0.5, 0.6, 0.7\}$, and the momentum over $\mu \in \{0.7, 0.8, 0.9\}$. 
For Delayed DiLoCo, the outer learning rate was selected by lowering by $\times4$ the optimal outer learning found for Eager DiLoCo~\citep{kale2025eager}.
Given the two-stage step mechanism of \ours{}, two inner learning rates $\eta_{in}$ and $\tilde \eta_{in}$ as well as two outer learning rates $\eta_{out}$ and $\tilde \eta_{out}$ should be all tuned. For simplicity, we set $\eta_{in}=\tilde \eta_{in}$ and $\eta_{out} = \tilde \eta_{out}$.
Table~\ref{tab:hyperparams_scaling} summarizes the values that were selected for each method.

\begin{table}[H]
\centering
\small
\caption{
\textbf{Selected hyperparameters across model sizes.}
}
\label{tab:hyperparams_scaling}

\setlength{\tabcolsep}{6pt}
\renewcommand{\arraystretch}{1.15}

\begin{tabular}{@{}lccccccccc@{}}
\toprule
\multirow{2}{*}{\textbf{Method}}
& \multicolumn{3}{c}{\textbf{(a) 134M model}}
& \multicolumn{3}{c}{\textbf{(b) 325M model}}
& \multicolumn{3}{c}{\modif{\textbf{(c) 500M model}}} \\
\cmidrule(lr){2-4}
\cmidrule(lr){5-7}
\cmidrule(lr){8-10}
& $\eta_{\mathrm{in}}$
& $\eta_{\mathrm{out}}$
& $\mu$
& $\eta_{\mathrm{in}}$
& $\eta_{\mathrm{out}}$
& $\mu$
& \modif{$\eta_{\mathrm{in}}$}
& \modif{$\eta_{\mathrm{out}}$}
& \modif{$\mu$} \\
\midrule
DDP
& $5\mathrm{e}{-3}$ & --- & ---
& $2\mathrm{e}{-3}$ & --- & ---
& \modif{$3\mathrm{e}{-3}$} & \modif{---} & \modif{---} \\
\noalign{\vskip 2pt}
\hdashline
\noalign{\vskip 2pt}
DiLoCo
& $1\mathrm{e}{-3}$ & $0.7$ & $0.9$
& $7\mathrm{e}{-4}$ & $0.7$ & $0.9$
& \modif{$8\mathrm{e}{-4}$} & \modif{$0.7$} & \modif{$0.9$} \\
\ours{}
& $2\mathrm{e}{-3}$ & $0.7$ & $0.9$
& \modif{$7\mathrm{e}{-4}$} & $0.7$ & $0.9$
& \modif{$2\mathrm{e}{-3}$} & \modif{$0.7$} & \modif{$0.9$} \\
Eager DiLoCo
& $2\mathrm{e}{-3}$ & $0.4$ & $0.9$
& \modif{$7\mathrm{e}{-4}$} & $0.4$ & $0.9$
& \modif{$8\mathrm{e}{-4}$} & \modif{$0.4$} & \modif{$0.9$} \\
Delayed DiLoCo
& $2\mathrm{e}{-3}$ & $0.1$ & $0.9$
& $8\mathrm{e}{-4}$ & $0.1$ & $0.9$
& \modif{$8\mathrm{e}{-4}$} & \modif{$0.1$} & \modif{$0.9$} \\
\bottomrule
\end{tabular}
\end{table}

\subsection{Implementation}\label{app:impl}

All experiments were conducted using PyTorch 2.13.0~\citep{paszke2019pytorch} with CUDA 12.6, cuDNN 9.10.2, building on the TorchTitan~\citep{liang2025torchtitan} and TorchFT codebases. Replica communication relies on NCCL 2.29.3 collectives~\citep{jeaugey2017nccl}, launched asynchronously to overlap communication with local computation. Models are trained in mixed precision~\citep{micikevicius2018mixed,kalamkar2019bfloat16} on NVIDIA A100 and H100 GPUs.

\section{Proofs}

\label{app:proofs}

This section collects the proofs of the main technical results.

\subsection{Auxiliary lemmas}

\begin{lemma}[Gradient Mismatch]\label{lemma:gradient-mismatch}
Let \(f\) be \(L\)-smooth. Fix \(H\ge 1\) and \(0<\beta L\le 1\). Define, for every replica \(i=1,\dots,K\), the local deterministic trajectory
\[
\theta_{t,0}^i=\theta_t,
\qquad
\theta_{t,h+1}^i
=
\theta_{t,h}^i-\frac{\beta}{H}\nabla f(\theta_{t,h}^i),
\qquad
h=0,\dots,H-1,
\]
and the accumulated local updates
\[
g_t^i:=\theta_t-\theta_{t,H}^i.
\]
Then
\begin{align*}
\bigl\|g_t^i-\beta\nabla f(\theta_t)\bigr\|
&\le
\frac{
\left(1+\frac{\beta L}{H}\right)^H-1-\beta L
}{L}
\|\nabla f(\theta_t)\| \\
&\le
\beta^2L\|\nabla f(\theta_t)\|.
\end{align*}
\end{lemma}

\begin{proof}
First note that
\[
g_t^i
=
\frac{\beta}{H}\sum_{h=0}^{H-1}\nabla f(\theta_{t,h}^i).
\]
Hence
\[
g_t^i-\beta\nabla f(\theta_t)
=
\frac{\beta}{H}\sum_{h=0}^{H-1}
\bigl(\nabla f(\theta_{t,h}^i)-\nabla f(\theta_t)\bigr).
\]
By \(L\)-smoothness,
\[
\bigl\|g_t^i-\beta\nabla f(\theta_t)\bigr\|
\le
\frac{\beta L}{H}
\sum_{h=0}^{H-1}\|\theta_{t,h}^i-\theta_t\|.
\]
We claim that
\[
\|\theta_{t,h}^i-\theta_t\|
\le
\frac{
\left(1+\frac{\beta L}{H}\right)^h-1
}{L}
\|\nabla f(\theta_t)\|.
\]
Indeed, using the update rule and \(L\)-smoothness,
\[
\|\theta_{t,h+1}^i-\theta_t\|
\le
\left(1+\frac{\beta L}{H}\right)
\|\theta_{t,h}^i-\theta_t\|
+
\frac{\beta}{H}\|\nabla f(\theta_t)\|.
\]
Since \(\theta_{t,0}^i=\theta_t\), iterating this recursion gives the claim.
Therefore,
\begin{align*}
\bigl\|g_t^i-\beta\nabla f(\theta_t)\bigr\|
&\le
\frac{\beta}{H}
\sum_{h=0}^{H-1}
\left[
\left(1+\frac{\beta L}{H}\right)^h-1
\right]
\|\nabla f(\theta_t)\| \\
&=
\frac{
\left(1+\frac{\beta L}{H}\right)^H-1-\beta L
}{L}
\|\nabla f(\theta_t)\|.
\end{align*}

Finally, since \(0<\beta L\le 1\),
\[
\left(1+\frac{\beta L}{H}\right)^H
\le e^{\beta L}
\]
and
\[
e^x-1-x\le x^2,
\qquad 0\le x \le 1.
\]
Thus
\[
\left(1+\frac{\beta L}{H}\right)^H-1-\beta L
\le
(\beta L)^2,
\]
which gives
\[
\bigl\|g_t^i-\beta\nabla f(\theta_t)\bigr\|
\le
\beta^2L\|\nabla f(\theta_t)\|.
\]
\end{proof}
\begin{lemma}[Expected Stochastic Gradient Mismatch]
\label{lemma:gradient-mismatch-sto}
Let \(f:\mathbb R^d\to\mathbb R\) be \(L\)-smooth. Let
\(\mathcal F_{t,h}\) be the filtration generated by the randomness up to
the \(h\)-th local step of outer iteration \(t\). Assume that:
\begin{enumerate}[leftmargin=*,labelsep=0.5em]
\item[(i)] The stochastic gradient is unbiased:
\[
\mathbb{E}\left[
\nabla F(\theta_{t,h}^i;\xi_{t,h}^i)
\mid
\mathcal{F}_{t,h}
\right]
=
\nabla f(\theta_{t,h}^i).
\]

\item[(ii)] The conditional variance is uniformly bounded by \(\sigma^2\):
\[
\mathbb{E}\left[
\left\|
\nabla F(\theta_{t,h}^i;\xi_{t,h}^i)
-
\nabla f(\theta_{t,h}^i)
\right\|^2
\mid
\mathcal{F}_{t,h}
\right]
\le
\sigma^2.
\]
\end{enumerate}

Fix \(H\ge 1\) and \(0<\beta L\le 1\). For every replica
\(i=1,\dots,K\), define the local stochastic trajectory by
\[
\theta_{t,0}^i=\theta_t,
\qquad
\theta_{t,h+1}^i
=
\theta_{t,h}^i
-
\frac{\beta}{H}
\nabla F(\theta_{t,h}^i;\xi_{t,h}^i),
\qquad
h=0,\dots,H-1.
\]
Define the accumulated local update
\[
g_t^i
:=
\theta_t-\theta_{t,H}^i
=
\frac{\beta}{H}
\sum_{h=0}^{H-1}
\nabla F(\theta_{t,h}^i;\xi_{t,h}^i).
\]
Then, with
\[
q:=1+\frac{\beta L}{H},
\]
we have
\begin{align*}
\left\|
\mathbb E[g_t^i]
-
\beta\nabla f(\theta_t)
\right\|^2
&\le
\frac{2\beta^2}{H}
\sum_{h=0}^{H-1}
(q^h-1)^2
\|\nabla f(\theta_t)\|^2
\\
&\quad+
\frac{2\beta^3L\sigma^2}{H^2}
\sum_{h=0}^{H-1}
\frac{q^{2h}-1}
{2+\frac{\beta L}{H}}.
\end{align*}
Consequently,
\[
\left\|
\mathbb E[g_t^i]
-
\beta\nabla f(\theta_t)
\right\|^2
\le
8\beta^4L^2
\|\nabla f(\theta_t)\|^2
+
8\frac{\beta^3L\sigma^2}{H}.
\]
\end{lemma}
\begin{proof}
Throughout the proof, all expectations are taken with respect to the
local stochastic-gradient randomness, conditional on the current
iterate $\theta_t$.

Define the stochastic gradient noise
\[
\varepsilon_{t,h}^i
:=
\nabla F(\theta_{t,h}^i;\xi_{t,h}^i)
-
\nabla f(\theta_{t,h}^i).
\]
By assumptions (i)--(ii),
\[
\mathbb E[\varepsilon_{t,h}^i\mid \mathcal F_{t,h}] = 0,
\qquad
\mathbb E[\|\varepsilon_{t,h}^i\|^2\mid \mathcal F_{t,h}]
\le \sigma^2.
\]
Also define
\[
\Delta_{t,h}^i
:=
\theta_{t,h}^i-\theta_t.
\]
Since $\theta_{t,0}^i=\theta_t$, we have
\[
\Delta_{t,0}^i=0.
\]
Moreover, using the local update rule,
\[
\Delta_{t,h+1}^i
=
\Delta_{t,h}^i
-
\frac{\beta}{H}
\nabla f(\theta_{t,h}^i)
-
\frac{\beta}{H}
\varepsilon_{t,h}^i.
\]
Adding and subtracting $\nabla f(\theta_t)$ gives
\[
\Delta_{t,h+1}^i
=
\Delta_{t,h}^i
-
\frac{\beta}{H}
\bigl(
\nabla f(\theta_{t,h}^i)-\nabla f(\theta_t)
\bigr)
-
\frac{\beta}{H}\nabla f(\theta_t)
-
\frac{\beta}{H}\varepsilon_{t,h}^i.
\]
Let
\[
q:=1+\frac{\beta L}{H}.
\]
Introduce the noiseless comparison sequence
$\bar\Delta_{t,h}^i$, defined by
\[
\bar\Delta_{t,0}^i=0,
\]
and
\[
\bar\Delta_{t,h+1}^i
=
\bar\Delta_{t,h}^i
-
\frac{\beta}{H}
\bigl(
\nabla f(\theta_t+\bar\Delta_{t,h}^i)
-
\nabla f(\theta_t)
\bigr)
-
\frac{\beta}{H}\nabla f(\theta_t).
\]
By $L$-smoothness,
\[
\|\bar\Delta_{t,h+1}^i\|
\le
q\|\bar\Delta_{t,h}^i\|
+
\frac{\beta}{H}\|\nabla f(\theta_t)\|.
\]
Iterating this inequality yields
\[
\|\bar\Delta_{t,h}^i\|
\le
\frac{\|\nabla f(\theta_t)\|}{L}
(q^h-1).
\]
Now consider
\[
\Delta_{t,h}^i-\bar\Delta_{t,h}^i.
\]
Using the two recursions above and $L$-smoothness,
\[
\mathbb E\!\left[
\|\Delta_{t,h+1}^i-\bar\Delta_{t,h+1}^i\|^2
\mid
\mathcal F_{t,h}
\right]
\le
q^2
\|\Delta_{t,h}^i-\bar\Delta_{t,h}^i\|^2
+
\frac{\beta^2\sigma^2}{H^2}.
\]
Iterating from
$\Delta_{t,0}^i-\bar\Delta_{t,0}^i=0$ gives
\[
\mathbb E\!\left[
\|\Delta_{t,h}^i-\bar\Delta_{t,h}^i\|^2
\right]
\le
\frac{\beta^2\sigma^2}{H^2}
\sum_{j=0}^{h-1}q^{2j}.
\]
Since
\[
\sum_{j=0}^{h-1}q^{2j}
=
\frac{q^{2h}-1}{q^2-1}
\]
and
\[
q^2-1
=
\frac{\beta L}{H}
\left(
2+\frac{\beta L}{H}
\right),
\]
we obtain
\[
\mathbb E\!\left[
\|\Delta_{t,h}^i-\bar\Delta_{t,h}^i\|^2
\right]
\le
\frac{\beta\sigma^2}{HL}
\frac{q^{2h}-1}
{2+\frac{\beta L}{H}}.
\]
Using
\[
\Delta_{t,h}^i
=
\bar\Delta_{t,h}^i
+
\bigl(
\Delta_{t,h}^i-\bar\Delta_{t,h}^i
\bigr)
\]
and
\[
\|a+b\|^2\le 2\|a\|^2+2\|b\|^2,
\]
we get
\[
\mathbb E\!\left[
\|\Delta_{t,h}^i\|^2
\right]
\le
\frac{2}{L^2}
(q^h-1)^2
\|\nabla f(\theta_t)\|^2
+
\frac{2\beta\sigma^2}{HL}
\frac{q^{2h}-1}
{2+\frac{\beta L}{H}}.
\]
Now, by definition,
\[
g_t^i
=
\frac{\beta}{H}
\sum_{h=0}^{H-1}
\nabla F(\theta_{t,h}^i;\xi_{t,h}^i).
\]
By the tower property and unbiasedness of the stochastic gradient,
\[
\mathbb E[g_t^i]
=
\frac{\beta}{H}
\sum_{h=0}^{H-1}
\mathbb E[
\nabla f(\theta_{t,h}^i)
].
\]
Therefore,
\[
\mathbb E[g_t^i]
-
\beta\nabla f(\theta_t)
=
\frac{\beta}{H}
\sum_{h=0}^{H-1}
\mathbb E\!\left[
\nabla f(\theta_{t,h}^i)
-
\nabla f(\theta_t)
\right].
\]
By Jensen's inequality, Cauchy--Schwarz, and $L$-smoothness,
\begin{align*}
\left\|
\mathbb E[g_t^i]
-
\beta\nabla f(\theta_t)
\right\|^2
&=
\left\|
\frac{\beta}{H}
\sum_{h=0}^{H-1}
\mathbb E\!\left[
\nabla f(\theta_{t,h}^i)
-
\nabla f(\theta_t)
\right]
\right\|^2
\\
&\le
\frac{\beta^2}{H}
\sum_{h=0}^{H-1}
\left\|
\mathbb E\!\left[
\nabla f(\theta_{t,h}^i)
-
\nabla f(\theta_t)
\right]
\right\|^2
\\
&\le
\frac{\beta^2}{H}
\sum_{h=0}^{H-1}
\mathbb E\!\left[
\|
\nabla f(\theta_{t,h}^i)
-
\nabla f(\theta_t)
\|^2
\right]
\\
&\le
\frac{\beta^2L^2}{H}
\sum_{h=0}^{H-1}
\mathbb E\!\left[
\|\Delta_{t,h}^i\|^2
\right].
\end{align*}
Substituting the previous estimate gives
\begin{align*}
\left\|
\mathbb E[g_t^i]
-
\beta\nabla f(\theta_t)
\right\|^2
&\le
\frac{2\beta^2}{H}
\sum_{h=0}^{H-1}
(q^h-1)^2
\|\nabla f(\theta_t)\|^2
\\
&\quad+
\frac{2\beta^3L\sigma^2}{H^2}
\sum_{h=0}^{H-1}
\frac{q^{2h}-1}
{2+\frac{\beta L}{H}}.
\end{align*}
Since $h\le H$ and $\beta L\le 1$,
\[
q^h
=
\left(
1+\frac{\beta L}{H}
\right)^h
\le
e^{\beta L}.
\]
Moreover,
\[
e^{\beta L}-1
\le
(e-1)\beta L,
\]
so
\[
q^h-1
\le
(e-1)\beta L.
\]
Also,
\[
q^{2h}
\le
e^{2\beta L}
\le
e^2,
\]
and therefore
\[
\frac{q^{2h}-1}
{2+\frac{\beta L}{H}}
\le
\frac{e^2-1}{2}.
\]
Consequently,
\begin{align*}
\left\|
\mathbb E[g_t^i]
-
\beta\nabla f(\theta_t)
\right\|^2
&\le
2(e-1)^2
\beta^4L^2
\|\nabla f(\theta_t)\|^2
\\
&\quad+
(e^2-1)
\frac{\beta^3L\sigma^2}{H}.
\end{align*}
Thus,
\[
\left\|
\mathbb E[g_t^i]
-
\beta\nabla f(\theta_t)
\right\|^2
\le
2(e-1)^2
\beta^4L^2
\|\nabla f(\theta_t)\|^2
+
(e^2-1)
\frac{\beta^3L\sigma^2}{H}.
\]
\end{proof}

\begin{lemma}[Averaged Stochastic Gradient Mismatch]
\label{lemma:replica-averaging}
Suppose that the assumptions and notation of
Lemma~\ref{lemma:gradient-mismatch-sto} hold. Set
\(\mathcal F_t:=\mathcal F_{t,0}\). Moreover, assume that, conditionally
on \(\mathcal F_t\), the local sample sequences
\[
(\xi_{t,0}^i,\ldots,\xi_{t,H-1}^i),
\qquad i=1,\ldots,K,
\]
are i.i.d. across replicas.

Let
\[
g_t:=\frac1K\sum_{i=1}^K g_t^i.
\]
Then
\[
\left\|
\mathbb E\!\left[g_t\middle|\mathcal F_t\right]
-
\beta\nabla f(\theta_t)
\right\|^2
\le
8\beta^4L^2
\|\nabla f(\theta_t)\|^2
+
8\frac{\beta^3L\sigma^2}{H}.
\]
\end{lemma}

\begin{proof}
By linearity of conditional expectation,
\[
\mathbb E\!\left[g_t\middle|\mathcal F_t\right]
=
\frac1K
\sum_{i=1}^K
\mathbb E\!\left[g_t^i\middle|\mathcal F_t\right].
\]
Hence,
\begin{align*}
\mathbb E\!\left[g_t\middle|\mathcal F_t\right]
-
\beta\nabla f(\theta_t)
&=
\frac1K
\sum_{i=1}^K
\left(
\mathbb E\!\left[g_t^i\middle|\mathcal F_t\right]
-
\beta\nabla f(\theta_t)
\right).
\end{align*}
Therefore, by Jensen's inequality,
\begin{align*}
\left\|
\mathbb E\!\left[g_t\middle|\mathcal F_t\right]
-
\beta\nabla f(\theta_t)
\right\|^2
&\le
\frac1K
\sum_{i=1}^K
\left\|
\mathbb E\!\left[g_t^i\middle|\mathcal F_t\right]
-
\beta\nabla f(\theta_t)
\right\|^2.
\end{align*}
Since the replicas are conditionally identically distributed given
\(\mathcal F_t\), the terms in the sum are identical. Thus,
\[
\left\|
\mathbb E\!\left[g_t\middle|\mathcal F_t\right]
-
\beta\nabla f(\theta_t)
\right\|^2
=
\left\|
\mathbb E\!\left[g_t^i\middle|\mathcal F_t\right]
-
\beta\nabla f(\theta_t)
\right\|^2.
\]
Applying Lemma~\ref{lemma:gradient-mismatch-sto} yields
\[
\left\|
\mathbb E\!\left[g_t\middle|\mathcal F_t\right]
-
\beta\nabla f(\theta_t)
\right\|^2
\le
8\beta^4L^2
\|\nabla f(\theta_t)\|^2
+
8\frac{\beta^3L\sigma^2}{H}.
\]
\end{proof}

\begin{lemma}[Conditional Variance of Averaged Updates]
\label{lemma:gradient-second-moment}
Suppose that the assumptions and notation of
Lemma~\ref{lemma:replica-averaging} hold.
Then
\[
\boxed{
\mathbb E\left[
\left\|
g_t-\mathbb E[g_t\mid\mathcal F_t]
\right\|^2
\mid\mathcal F_t
\right]
\le
\frac{e^2\beta^2\sigma^2}{KH}.
}
\]
\end{lemma}

\begin{proof}
Let $\bar\theta_{t,h}$ denote the deterministic
comparison trajectory initialized at $\theta_t$,
defined by
\[
\bar\theta_{t,0}=\theta_t,
\qquad
\bar\theta_{t,h+1}
=
\bar\theta_{t,h}
-\frac{\beta}{H}
\nabla f(\bar\theta_{t,h}),
\]
and set
\[
\bar g_t:=\theta_t-\bar\theta_{t,H}.
\]

Define
\[
d_{t,h}^i
:=
\theta_{t,h}^i-\bar\theta_{t,h},
\qquad
q:=1+\frac{\beta L}{H}.
\]
By $L$-smoothness, conditional unbiasedness,
and the variance bound,
\begin{align*}
\mathbb E\left[
\|d_{t,h+1}^i\|^2
\mid\mathcal F_{t,h}
\right]
&\le
q^2\|d_{t,h}^i\|^2
+\frac{\beta^2\sigma^2}{H^2}.
\end{align*}
Since $d_{t,0}^i=0$, iterating yields
\begin{align*}
\mathbb E\left[
\|d_{t,H}^i\|^2
\mid\mathcal F_t
\right]
&\le
\frac{\beta^2\sigma^2}{H^2}
\sum_{h=0}^{H-1}q^{2h}
\\
&\le
\frac{e^{2\beta L}\beta^2\sigma^2}{H}
\\
&\le
\frac{e^2\beta^2\sigma^2}{H},
\end{align*}
where we used $\beta L\le1$.

Since $g_t^i-\bar g_t=-d_{t,H}^i$
and $\bar g_t$ is $\mathcal F_t$-measurable,
the conditional mean minimizes the expected
squared deviation, so
\begin{align*}
&\mathbb E\left[
\left\|
g_t^i-\mathbb E[g_t^i\mid\mathcal F_t]
\right\|^2
\mid\mathcal F_t
\right]
\\
&\le
\mathbb E\left[
\|g_t^i-\bar g_t\|^2
\mid\mathcal F_t
\right]
\\
&\le
\frac{e^2\beta^2\sigma^2}{H}.
\end{align*}

Finally, by conditional independence
across replicas and
$g_t=\frac1K\sum_{i=1}^K g_t^i$,
\begin{align*}
&\mathbb E\left[
\left\|
g_t-\mathbb E[g_t\mid\mathcal F_t]
\right\|^2
\mid\mathcal F_t
\right]
\\
&=
\frac1{K^2}\sum_{i=1}^K
\mathbb E\left[
\left\|
g_t^i-\mathbb E[g_t^i\mid\mathcal F_t]
\right\|^2
\mid\mathcal F_t
\right]
\\
&\le
\frac{e^2\beta^2\sigma^2}{KH}.
\end{align*}
\end{proof}

\subsection{Main theorems}

\begin{apptheorem}[Deterministic Lyapunov Descent]
\label{theorem:lyapunov-descent}
Let \(f:\mathbb R^d\to\mathbb R\) be \(L\)-smooth
and bounded below by \(f^\star\).
Fix \(H\ge1\), set \(\eta=30\), and choose
\(\beta>0\) such that
\[
0<\beta L\le\frac{1}{30}.
\]
For every replica \(i=1,\dots,K\), define
the deterministic local trajectories
\[
\theta_{t,0}^i=\theta_t,
\qquad
\theta_{t,h+1}^i
=
\theta_{t,h}^i
-
\frac{\beta}{H}\nabla f(\theta_{t,h}^i),
\qquad
h=0,\dots,H-1,
\]
and
\[
\tilde\theta_{t,0}^i=\tilde\theta_t,
\qquad
\tilde\theta_{t,h+1}^i
=
\tilde\theta_{t,h}^i
-
\frac{\beta}{H}\nabla f(\tilde\theta_{t,h}^i),
\qquad
h=0,\dots,H-1.
\]
Define the accumulated local updates
\[
g_t^i:=\theta_t-\theta_{t,H}^i,
\qquad
\tilde g_t^i:=\tilde\theta_t-\tilde\theta_{t,H}^i,
\]
and their replica averages
\[
g_t=\frac1K\sum_{i=1}^K g_t^i,
\qquad
\tilde g_t=\frac1K\sum_{i=1}^K\tilde g_t^i.
\]
Consider the outer updates
\[
\theta_{t+1}
=
\theta_t-\frac{\eta}{2}(g_t+\tilde g_t),
\qquad
\tilde\theta_{t+1}
=
\theta_t-\eta\tilde g_t.
\]
Define the Lyapunov function
\[
\mathcal V_t
:=
f(\theta_t)-f^\star
+
\eta\beta L(f(\tilde\theta_t)-f^\star)
+
16L\|\theta_t-\tilde\theta_t\|^2.
\]
Then
\[
\boxed{
\mathcal V_{t+1}-\mathcal V_t
\le
-\frac{\eta\beta}{3}
\|\nabla f(\theta_t)\|^2.
}
\]
Consequently, for every \(T\ge1\),
\[
\frac1T\sum_{t=0}^{T-1}
\|\nabla f(\theta_t)\|^2
\le
\frac{3\mathcal V_0}{\eta\beta T}.
\]
Moreover, if \(\theta_0=\tilde\theta_0\), then
\[
\boxed{
\frac1T\sum_{t=0}^{T-1}
\|\nabla f(\theta_t)\|^2
\le
\frac{6\bigl(f(\theta_0)-f^\star\bigr)}
{\eta\beta T}.
}
\]
\end{apptheorem}

\begin{proof}
By \(L\)-smoothness and the update
\[
\theta_{t+1}
=
\theta_t-\frac{\eta}{2}(g_t+\tilde g_t),
\]
we have
\begin{align*}
f(\theta_{t+1})-f(\theta_t)
&\le
-\eta
\left\langle
\nabla f(\theta_t),
\frac{g_t+\tilde g_t}{2}
\right\rangle
+
\frac{L\eta^2}{2}
\left\|
\frac{g_t+\tilde g_t}{2}
\right\|^2 .
\end{align*}
Using the identity
\[
-\langle a,b\rangle
=
-\frac{\beta}{2}\|a\|^2
-\frac{1}{2\beta}\|b\|^2
+
\frac{1}{2\beta}\|b-\beta a\|^2,
\]
with
\[
a=\nabla f(\theta_t),
\qquad
b=\frac{g_t+\tilde g_t}{2},
\]
we obtain
\begin{align*}
f(\theta_{t+1})-f(\theta_t)
&\le
-\frac{\eta\beta}{2}\|\nabla f(\theta_t)\|^2
-
\frac{\eta}{2\beta}
\left(
1-\eta\beta L
\right)
\left\|
\frac{g_t+\tilde g_t}{2}
\right\|^2
\\
&\quad
+
\frac{\eta}{2\beta}
\left\|
\frac{g_t+\tilde g_t}{2}
-
\beta\nabla f(\theta_t)
\right\|^2 .
\end{align*}
Since \(\eta\beta L\le1\), the middle term is nonpositive and may be dropped.
Moreover,
\begin{align*}
\frac{g_t+\tilde g_t}{2}
-
\beta\nabla f(\theta_t)
&=
\frac12
\left(
g_t-\beta\nabla f(\theta_t)
\right)
+
\frac12
\left(
\tilde g_t-\beta\nabla f(\tilde\theta_t)
\right)
\\
&\quad
+
\frac{\beta}{2}
\left(
\nabla f(\tilde\theta_t)-\nabla f(\theta_t)
\right).
\end{align*}
Therefore, using \(\|a+b+c\|^2\le3\|a\|^2+3\|b\|^2+3\|c\|^2\) and
\(L\)-smoothness,
\begin{align*}
\left\|
\frac{g_t+\tilde g_t}{2}
-
\beta\nabla f(\theta_t)
\right\|^2
&\le
\frac34
\left\|
g_t-\beta\nabla f(\theta_t)
\right\|^2
+
\frac34
\left\|
\tilde g_t-\beta\nabla f(\tilde\theta_t)
\right\|^2
\\
&\quad
+
\frac{3\beta^2L^2}{4}
\|\theta_t-\tilde\theta_t\|^2 .
\end{align*}
Hence
\begin{align*}
f(\theta_{t+1})-f(\theta_t)
&\le
-\frac{\eta\beta}{2}\|\nabla f(\theta_t)\|^2
+
\frac{3\eta}{8\beta}
\left\|
g_t-\beta\nabla f(\theta_t)
\right\|^2
\\
&\quad
+
\frac{3\eta}{8\beta}
\left\|
\tilde g_t-\beta\nabla f(\tilde\theta_t)
\right\|^2
+
\frac{3\eta\beta L^2}{8}
\|\theta_t-\tilde\theta_t\|^2 .
\end{align*}
Next, by \(L\)-smoothness and the update
\[
\tilde\theta_{t+1}
=
\theta_t-\eta\tilde g_t,
\]
we have
\begin{align*}
\eta \beta L(f(\tilde\theta_{t+1})-f(\tilde\theta_t))
&\le
\eta \beta  L
\left\langle
\nabla f(\tilde\theta_t),
\theta_t-\tilde\theta_t-\eta\tilde g_t
\right\rangle
+
\frac{\eta \beta  L^2}{2}
\left\|
\theta_t-\tilde\theta_t-\eta\tilde g_t
\right\|^2
\\
&=
-\eta \beta  L
\left\langle
\nabla f(\tilde\theta_t),
\eta\tilde g_t-\theta_t+\tilde\theta_t
\right\rangle
+
\frac{\eta \beta  L^2}{2}
\left\|
\eta\tilde g_t-\theta_t+\tilde\theta_t
\right\|^2 .
\end{align*}
Using the identity
\[
-\langle a,b\rangle
=
-\frac{\gamma}{2}\|a\|^2
-\frac{1}{2\gamma}\|b\|^2
+
\frac{1}{2\gamma}\|b-\gamma a\|^2,
\]
with
\[
a= \eta\beta\nabla f(\tilde\theta_t),
\qquad
b=\eta\tilde g_t-\theta_t+\tilde\theta_t,
\qquad
\gamma=1,
\]
gives

\begin{align*}
\eta\beta L(f(\tilde\theta_{t+1})-f(\tilde\theta_t))
&\le
-\frac{\eta^2\beta^2 L}{2}
\|\nabla f(\tilde\theta_t)\|^2
\\
&\quad
-\frac{L}{2}(1-\eta\beta L)
\left\|
\eta\tilde g_t-\theta_t+\tilde\theta_t
\right\|^2
\\
&\quad
+\frac{L}{2}
\left\|
\eta\tilde g_t-\theta_t+\tilde\theta_t
-\eta\beta\nabla f(\tilde\theta_t)
\right\|^2.
\end{align*}

Since \(\eta\beta L\le1\), the middle term is nonpositive and may be dropped.
Furthermore,
\begin{align*}
\eta\tilde g_t-\theta_t+\tilde\theta_t
-
\eta\beta\nabla f(\tilde\theta_t)
=
\eta
\left(
\tilde g_t-\beta\nabla f(\tilde\theta_t)
\right)
-
(\theta_t-\tilde\theta_t).
\end{align*}
Thus, by \(\|a-b\|^2\le2\|a\|^2+2\|b\|^2\),
\begin{align*}
\left\|
\eta\tilde g_t-\theta_t+\tilde\theta_t
-
\eta\beta\nabla f(\tilde\theta_t)
\right\|^2
&\le
2\eta^2
\left\|
\tilde g_t-\beta\nabla f(\tilde\theta_t)
\right\|^2
+
2\|\theta_t-\tilde\theta_t\|^2 .
\end{align*}
Therefore,
\begin{align*}
\eta\beta L
\left(f(\tilde\theta_{t+1})-f(\tilde\theta_t)\right)
&\le
-\frac{\eta^2\beta^2 L}{2}
\|\nabla f(\tilde\theta_t)\|^2
\\
&\quad
+\eta^2 L
\left\|
\tilde g_t-\beta\nabla f(\tilde\theta_t)
\right\|^2
\\
&\quad
+L\|\theta_t-\tilde\theta_t\|^2.
\end{align*}

Finally, using the two outer updates,
\begin{align*}
\theta_{t+1}-\tilde\theta_{t+1}
&=
\theta_t-\frac{\eta}{2}(g_t+\tilde g_t)
-
\left(
\theta_t-\eta\tilde g_t
\right)
\\
&=
\frac{\eta}{2}(\tilde g_t-g_t).
\end{align*}
Hence
\begin{align*}
\|\theta_{t+1}-\tilde\theta_{t+1}\|^2
-
\|\theta_t-\tilde\theta_t\|^2
&=
-\|\theta_t-\tilde\theta_t\|^2
+
\frac{\eta^2}{4}
\|g_t-\tilde g_t\|^2 .
\end{align*}
Moreover,
\begin{align*}
g_t-\tilde g_t
&=
\left(
g_t-\beta\nabla f(\theta_t)
\right)
-
\left(
\tilde g_t-\beta\nabla f(\tilde\theta_t)
\right)
\\
&\quad
+
\beta
\left(
\nabla f(\theta_t)-\nabla f(\tilde\theta_t)
\right).
\end{align*}
Therefore, using \(\|a+b+c\|^2\le3\|a\|^2+3\|b\|^2+3\|c\|^2\) and
\(L\)-smoothness,
\begin{align*}
\|g_t-\tilde g_t\|^2
&\le
3
\left\|
g_t-\beta\nabla f(\theta_t)
\right\|^2
+
3
\left\|
\tilde g_t-\beta\nabla f(\tilde\theta_t)
\right\|^2
\\
&\quad
+
3\beta^2L^2
\|\theta_t-\tilde\theta_t\|^2 .
\end{align*}
If $\kappa=16L$, let's pursue:
\begin{align*}
\kappa
\left(
\|\theta_{t+1}-\tilde\theta_{t+1}\|^2
-
\|\theta_t-\tilde\theta_t\|^2
\right)
&\le
-\kappa
\left(
1-\frac{3\eta^2\beta^2L^2}{4}
\right)
\|\theta_t-\tilde\theta_t\|^2
\\
&\quad
+
\frac{3\kappa\eta^2}{4}
\left\|
g_t-\beta\nabla f(\theta_t)
\right\|^2
\\
&\quad
+
\frac{3\kappa\eta^2}{4}
\left\|
\tilde g_t-\beta\nabla f(\tilde\theta_t)
\right\|^2 .
\end{align*}
Adding the three estimates gives

\begin{align*}
\mathcal V_{t+1}-\mathcal V_t
&\le
-\frac{\eta\beta}{2}
\|\nabla f(\theta_t)\|^2
-\frac{\eta^2\beta^2 L}{2}
\|\nabla f(\tilde\theta_t)\|^2
\\
&\quad
+\left(
\frac{3\eta}{8\beta}
+\frac{3\kappa\eta^2}{4}
\right)
\left\|
g_t-\beta\nabla f(\theta_t)
\right\|^2
\\
&\quad
+\left(
\frac{3\eta}{8\beta}
+\eta^2 L
+\frac{3\kappa\eta^2}{4}
\right)
\left\|
\tilde g_t-\beta\nabla f(\tilde\theta_t)
\right\|^2
\\
&\quad
-\left[
\kappa\left(
1-\frac{3\eta^2\beta^2 L^2}{4}
\right)
-L
-\frac{3\eta\beta L^2}{8}
\right]
\|\theta_t-\tilde\theta_t\|^2.
\end{align*}

For \(\kappa=16L\), the coefficient of
\(\|\theta_t-\tilde\theta_t\|^2\) satisfies
\begin{align*}
&-\kappa\left(
1-\frac{3\eta^2\beta^2L^2}{4}
\right)
+L+\frac{3\eta\beta L^2}{8}
\\
&=
-15L+12\eta^2\beta^2L^3
+\frac{3\eta\beta L^2}{8}
\\
&\le
-15L+12L+\frac{3L}{8}
\\
&=
-\frac{21L}{8}<0,
\end{align*}
where we used \(\eta\beta L\le 1\). Moreover, for \(\kappa=16L\), the coefficients
of the error terms satisfy
\begin{align*}
\frac{3\eta}{8\beta}
+\frac{3\kappa\eta^2}{4}
&=
\frac{3\eta}{8\beta}
+12\eta^2 L
\\
&\le
\frac{99\eta}{8\beta},
\end{align*}
and
\begin{align*}
\frac{3\eta}{8\beta}
+\eta^2L
+\frac{3\kappa\eta^2}{4}
&=
\frac{3\eta}{8\beta}
+13\eta^2L
\\
&\le
\frac{107\eta}{8\beta},
\end{align*}
where we used \(\eta\beta L\le1\). Consequently,
\begin{align*}
\mathcal V_{t+1}-\mathcal V_t
&\le
-\frac{\eta\beta}{2}
\|\nabla f(\theta_t)\|^2
-\frac{\eta^2\beta^2L}{2}
\|\nabla f(\tilde\theta_t)\|^2
\\
&\quad
+\frac{99\eta}{8\beta}
\left\|
g_t-\beta\nabla f(\theta_t)
\right\|^2
\\
&\quad
+\frac{107\eta}{8\beta}
\left\|
\tilde g_t-\beta\nabla f(\tilde\theta_t)
\right\|^2
\\
&\quad
-\frac{21L}{8}
\|\theta_t-\tilde\theta_t\|^2.
\end{align*}

By Lemma~\ref{lemma:gradient-mismatch}, for every replica \(i\),
\[
\|g_t^i-\beta\nabla f(\theta_t)\|
\le
\beta^2L\|\nabla f(\theta_t)\|,
\]
and
\[
\|\tilde g_t^i-\beta\nabla f(\tilde\theta_t)\|
\le
\beta^2L\|\nabla f(\tilde\theta_t)\|.
\]
Jensen's inequality for the replica averages then gives
\[
\|g_t-\beta\nabla f(\theta_t)\|^2
\le
\beta^4L^2\|\nabla f(\theta_t)\|^2
\]
and
\[
\|\tilde g_t-\beta\nabla f(\tilde\theta_t)\|^2
\le
\beta^4L^2\|\nabla f(\tilde\theta_t)\|^2.
\]
Substituting these estimates yields

\begin{align*}
\mathcal V_{t+1}-\mathcal V_t
\le{}&
-\frac{\eta\beta}{2}
\|\nabla f(\theta_t)\|^2
-\frac{\eta^2\beta^2L}{2}
\|\nabla f(\tilde\theta_t)\|^2
\\
&+
\frac{99\eta\beta^3L^2}{8}
\|\nabla f(\theta_t)\|^2
\\
&+
\frac{107\eta\beta^3L^2}{8}
\|\nabla f(\tilde\theta_t)\|^2
\\
&-
\frac{21L}{8}
\|\theta_t-\tilde\theta_t\|^2
\\
={}&
-\frac{\eta\beta}{2}
\left(
1-\frac{99}{4}\beta^2L^2
\right)
\|\nabla f(\theta_t)\|^2
\\
&-
\frac{\eta^2\beta^2L}{2}
\left(
1-\frac{107\beta L}{4\eta}
\right)
\|\nabla f(\tilde\theta_t)\|^2
\\
&-
\frac{21L}{8}
\|\theta_t-\tilde\theta_t\|^2.
\end{align*}

Fix $\eta=30$ and assume $\beta L\le 1/30$.
Then
\begin{align*}
1-\frac{99}{4}\beta^2L^2
&\ge 1-\frac{99}{3600}
\ge \frac23,
\\
1-\frac{107\beta L}{4\eta}
&\ge 1-\frac{107}{3600}
\ge \frac23.
\end{align*}
Therefore, dropping the nonpositive terms,
\begin{align*}
\mathcal V_{t+1}-\mathcal V_t
&\le
-\frac{\eta\beta}{3}
\|\nabla f(\theta_t)\|^2.
\end{align*}

Summing from $t=0$ to $T-1$ and using
$\mathcal V_T\ge0$, we obtain
\[
\frac{\eta\beta}{3}
\sum_{t=0}^{T-1}
\|\nabla f(\theta_t)\|^2
\le \mathcal V_0.
\]
Dividing by $\eta\beta T/3$ gives
\[
\frac1T\sum_{t=0}^{T-1}
\|\nabla f(\theta_t)\|^2
\le
\frac{3\mathcal V_0}{\eta\beta T}.
\]

Moreover, if $\theta_0=\tilde\theta_0$, then
\begin{align*}
\mathcal V_0
&=
(1+\eta\beta L)
\bigl(f(\theta_0)-f^\star\bigr)
\\
&\le
2\bigl(f(\theta_0)-f^\star\bigr),
\end{align*}
where we used $\eta\beta L\le1$.
Hence,
\[
\boxed{
\frac1T\sum_{t=0}^{T-1}
\|\nabla f(\theta_t)\|^2
\le
\frac{6\bigl(f(\theta_0)-f^\star\bigr)}
{\eta\beta T}.
}
\]
\end{proof}

\begin{apptheorem}[Stochastic Lyapunov Descent]
\label{theorem:stochastic-lyapunov-descent}
Let \(f:\mathbb R^d\to\mathbb R\) be \(L\)-smooth
and bounded below by \(f^\star\).
Suppose that the assumptions of
Lemmas~\ref{lemma:gradient-mismatch-sto},
\ref{lemma:replica-averaging}, and
\ref{lemma:gradient-second-moment} hold.
Set \(\mathcal F_t:=\mathcal F_{t,0}\).
Fix \(H\ge1\), set \(\eta=30\), and choose
\(\beta>0\) such that
\[
0<\beta L\le\frac{1}{30}.
\]
For every replica \(i=1,\dots,K\), define
the stochastic local trajectories
\[
\theta_{t,0}^i=\theta_t,
\qquad
\theta_{t,h+1}^i
=
\theta_{t,h}^i
-
\frac{\beta}{H}
\nabla F(\theta_{t,h}^i;\xi_{t,h}^i),
\qquad h=0,\dots,H-1,
\]
and
\[
\tilde\theta_{t,0}^i=\tilde\theta_t,
\qquad
\tilde\theta_{t,h+1}^i
=
\tilde\theta_{t,h}^i
-
\frac{\beta}{H}
\nabla F(\tilde\theta_{t,h}^i;\tilde\xi_{t,h}^i),
\qquad h=0,\dots,H-1.
\]
Define the accumulated local updates
\[
g_t^i:=\theta_t-\theta_{t,H}^i,
\qquad
\tilde g_t^i:=\tilde\theta_t-\tilde\theta_{t,H}^i,
\]
and their replica averages
\[
g_t=\frac1K\sum_{i=1}^K g_t^i,
\qquad
\tilde g_t=\frac1K\sum_{i=1}^K\tilde g_t^i.
\]
Consider the outer updates
\[
\theta_{t+1}
=
\theta_t-\frac{\eta}{2}(g_t+\tilde g_t),
\qquad
\tilde\theta_{t+1}
=
\theta_t-\eta\tilde g_t.
\]
If \(\theta_0=\tilde\theta_0\)
and the initialization is deterministic, then
\[
\boxed{
\begin{aligned}
\frac1T\sum_{t=0}^{T-1}
\mathbb E\left[
\|\nabla f(\theta_t)\|^2
\right]
&\le
\frac{
4(f(\theta_0)-f^\star)
}{15\beta T}
\\
&\quad+
\frac{824\beta L\sigma^2}{H}
+
\frac{18360\beta L\sigma^2}{KH}.
\end{aligned}
}
\]
\end{apptheorem}

\begin{proof}
First, we note that taking each terms of the Lyapunov function independently leads to:
\begin{align*}
&\mathbb E\left[
f(\theta_{t+1})-f(\theta_t)
\mid\mathcal F_t
\right]
\\
&\le
-\eta\left\langle
\nabla f(\theta_t),
\frac{
\mathbb E[g_t\mid\mathcal F_t]
+\mathbb E[\tilde g_t\mid\mathcal F_t]
}{2}
\right\rangle
\\
&\quad+
\frac{L\eta^2}{2}
\left\|
\frac{\mathbb{E}[g_t\mid\mathcal F_t]+\mathbb{E}[\tilde g_t\mid\mathcal F_t]}{2}
\right\|^2 .
\\
&\quad+
\frac{L\eta^2}{4}
\left(
\mathbb E\left[
\|g_t-\mathbb E[g_t\mid\mathcal F_t]\|^2
\mid\mathcal F_t
\right]
\right.
\\
&\qquad\qquad\left.
+
\mathbb E\left[
\|\tilde g_t-\mathbb E[\tilde g_t\mid\mathcal F_t]\|^2
\mid\mathcal F_t
\right]
\right).
\end{align*}

and

\begin{align*}
&\mathbb E\left[
\eta\beta L
(f(\tilde\theta_{t+1})-f(\tilde\theta_t))
\mid\mathcal F_t
\right]
\\
&\le
\eta\beta L
\left\langle
\nabla f(\tilde\theta_t),
\theta_t-\tilde\theta_t
-\eta\mathbb E[\tilde g_t\mid\mathcal F_t]
\right\rangle
\\
&\quad+
\frac{\eta\beta L^2}{2}
\left\|
\theta_t-\tilde\theta_t
-\eta\mathbb E[\tilde g_t\mid\mathcal F_t]
\right\|^2
\\
&\quad+
\frac{\eta^3\beta L^2}{2}
\mathbb E\left[
\left\|
\tilde g_t-
\mathbb E[\tilde g_t\mid\mathcal F_t]
\right\|^2
\mid\mathcal F_t
\right]
\end{align*}

and

\begin{align*}
&\mathbb E\left[
\|\theta_{t+1}-\tilde\theta_{t+1}\|^2
-
\|\theta_t-\tilde\theta_t\|^2
\mid\mathcal F_t
\right]
\\
&=
-\|\theta_t-\tilde\theta_t\|^2
+
\frac{\eta^2}{4}
\mathbb E\left[
\|g_t-\tilde g_t\|^2
\mid\mathcal F_t
\right]
\\
&\le
-\|\theta_t-\tilde\theta_t\|^2
+
\frac{\eta^2}{4}
\left\|
\mathbb E[g_t\mid\mathcal F_t]
-
\mathbb E[\tilde g_t\mid\mathcal F_t]
\right\|^2
\\
&\quad+
\frac{\eta^2}{2}
\mathbb E\left[
\|g_t-\mathbb E[g_t\mid\mathcal F_t]\|^2
+
\|\tilde g_t-\mathbb E[\tilde g_t\mid\mathcal F_t]\|^2
\mid\mathcal F_t
\right]
\end{align*}

Following the proof of
Theorem~\ref{theorem:lyapunov-descent},
the deterministic part of the argument
remains unchanged when replacing the
stochastic updates by their conditional means. We obtain
\begin{align*}
&\mathbb E\left[
\mathcal V_{t+1}-\mathcal V_t
\mid\mathcal F_t
\right]
\\
&\le
-\frac{\eta\beta}{2}
\|\nabla f(\theta_t)\|^2
-\frac{\eta^2\beta^2L}{2}
\|\nabla f(\tilde\theta_t)\|^2
\\
&\quad
+\frac{99\eta}{8\beta}
\left\|
\mathbb E[g_t\mid\mathcal F_t]
-\beta\nabla f(\theta_t)
\right\|^2
\\
&\quad
+\frac{107\eta}{8\beta}
\left\|
\mathbb E[\tilde g_t\mid\mathcal F_t]
-\beta\nabla f(\tilde\theta_t)
\right\|^2
\\
&\quad
-\frac{21L}{8}
\|\theta_t-\tilde\theta_t\|^2
\\
&\quad
+\underbrace{
\frac{L\eta^2}{2}
\frac{e^2\beta^2\sigma^2}{KH}
}_{\text{Main iterate}}
\\
&\quad
+\underbrace{
\frac{\eta^3\beta L^2}{2}
\frac{e^2\beta^2\sigma^2}{KH}
}_{\text{Auxiliary iterate}}
\\
&\quad
+\underbrace{
16L\eta^2
\frac{e^2\beta^2\sigma^2}{KH}
}_{\text{Discrepancy term}}
\\
&=
-\frac{\eta\beta}{2}
\|\nabla f(\theta_t)\|^2
-\frac{\eta^2\beta^2L}{2}
\|\nabla f(\tilde\theta_t)\|^2
\\
&\quad
+\frac{99\eta}{8\beta}
\left\|
\mathbb E[g_t\mid\mathcal F_t]
-\beta\nabla f(\theta_t)
\right\|^2
\\
&\quad
+\frac{107\eta}{8\beta}
\left\|
\mathbb E[\tilde g_t\mid\mathcal F_t]
-\beta\nabla f(\tilde\theta_t)
\right\|^2
\\
&\quad
-\frac{21L}{8}
\|\theta_t-\tilde\theta_t\|^2
\\
&\quad
+\frac{e^2\eta^2\beta^2L\sigma^2}{2KH}
(33+\eta\beta L).
\end{align*}

By Lemma~\ref{lemma:replica-averaging},
\[
\left\|
\mathbb E[g_t\mid\mathcal F_t]
-
\beta\nabla f(\theta_t)
\right\|^2
\le
2(e-1)^2\beta^4L^2
\|\nabla f(\theta_t)\|^2
+
(e^2-1)\frac{\beta^3L\sigma^2}{H},
\]
and similarly,
\[
\left\|
\mathbb E[\tilde g_t\mid\mathcal F_t]
-
\beta\nabla f(\tilde\theta_t)
\right\|^2
\le
2(e-1)^2\beta^4L^2
\|\nabla f(\tilde\theta_t)\|^2
+
(e^2-1)\frac{\beta^3L\sigma^2}{H}.
\]

Using
\[
(e-1)^2\le 8,
\qquad
e^2-1\le 8,
\]
we obtain
\[
\left\|
\mathbb E[g_t\mid\mathcal F_t]
-
\beta\nabla f(\theta_t)
\right\|^2
\le
16\beta^4L^2
\|\nabla f(\theta_t)\|^2
+
8\frac{\beta^3L\sigma^2}{H},
\]
and
\[
\left\|
\mathbb E[\tilde g_t\mid\mathcal F_t]
-
\beta\nabla f(\tilde\theta_t)
\right\|^2
\le
16\beta^4L^2
\|\nabla f(\tilde\theta_t)\|^2
+
8\frac{\beta^3L\sigma^2}{H}.
\]
Substituting these estimates gives

\begin{align*}
&\mathbb E\!\left[
\mathcal V_{t+1}-\mathcal V_t
\middle|\mathcal F_t
\right]
\\
&\le
-\frac{\eta\beta}{2}
\|\nabla f(\theta_t)\|^2
-\frac{\eta^2\beta^2L}{2}
\|\nabla f(\tilde\theta_t)\|^2
\\
&\quad
+198\eta\beta^3L^2
\|\nabla f(\theta_t)\|^2
\\
&\quad
+214\eta\beta^3L^2
\|\nabla f(\tilde\theta_t)\|^2
\\
&\quad
-\frac{21L}{8}
\|\theta_t-\tilde\theta_t\|^2
\\
&\quad
+\frac{206\eta\beta^2L\sigma^2}{H}
+\frac{e^2\eta^2\beta^2L\sigma^2}{2KH}
(33+\eta\beta L).
\\
&=
-\frac{\eta\beta}{2}
\left(1-396\beta^2L^2\right)
\|\nabla f(\theta_t)\|^2
\\
&\quad
-\frac{\eta^2\beta^2L}{2}
\left(1-\frac{428\beta L}{\eta}\right)
\|\nabla f(\tilde\theta_t)\|^2
\\
&\quad
-\frac{21L}{8}
\|\theta_t-\tilde\theta_t\|^2
\\
&\quad
+\frac{206\eta\beta^2L\sigma^2}{H}
+\frac{e^2\eta^2\beta^2L\sigma^2}{2KH}
(33+\eta\beta L).
\end{align*}

Fix $\eta=30$ and assume $0<\beta L\le 1/30$.
Then
\begin{align*}
1-396\beta^2L^2
&\ge
1-\frac{396}{900}
=\frac{14}{25}
\ge\frac12,
\\
1-\frac{428\beta L}{\eta}
&\ge
1-\frac{428}{900}
=\frac{118}{225}
\ge\frac12.
\end{align*}

Therefore, dropping the nonpositive terms
involving $\nabla f(\tilde\theta_t)$ and
$\theta_t-\tilde\theta_t$, we obtain
\begin{align*}
&\mathbb E\left[
\mathcal V_{t+1}-\mathcal V_t
\mid\mathcal F_t
\right]
\\
&\le
-\frac{15\beta}{2}
\|\nabla f(\theta_t)\|^2
+
\frac{6180\beta^2L\sigma^2}{H}
+
\frac{137700\beta^2L\sigma^2}{KH}.
\end{align*}

Taking total expectation and summing from
$t=0$ to $T-1$, and using
$\mathcal V_T\ge0$, yields
\begin{align*}
\frac{15\beta}{2}
\sum_{t=0}^{T-1}
\mathbb E\left[
\|\nabla f(\theta_t)\|^2
\right]
&\le
\mathbb E[\mathcal V_0]
\\
&\quad+
\frac{6180\beta^2L\sigma^2T}{H}
\\
&\quad+
\frac{137700\beta^2L\sigma^2T}{KH}.
\end{align*}

Dividing by $15\beta T/2$ gives
\begin{align*}
\frac1T\sum_{t=0}^{T-1}
\mathbb E\left[
\|\nabla f(\theta_t)\|^2
\right]
&\le
\frac{2\mathbb E[\mathcal V_0]}
{15\beta T}
\\
&\quad+
\frac{824\beta L\sigma^2}{H}
+
\frac{18360\beta L\sigma^2}{KH}.
\end{align*}

Moreover, if $\theta_0=\tilde\theta_0$
and the initialization is deterministic, then
\begin{align*}
\mathcal V_0
&=
(1+30\beta L)
\bigl(f(\theta_0)-f^\star\bigr)
\\
&\le
2\bigl(f(\theta_0)-f^\star\bigr),
\end{align*}
where we used $\beta L\le1/30$.

Consequently,
\[
\boxed{
\begin{aligned}
\frac1T\sum_{t=0}^{T-1}
\mathbb E\left[
\|\nabla f(\theta_t)\|^2
\right]
&\le
\frac{4(f(\theta_0)-f^\star)}
{15\beta T}
\\
&\quad+
\frac{824\beta L\sigma^2}{H}
+
\frac{18360\beta L\sigma^2}{KH}.
\end{aligned}
}
\]

\end{proof}

\section{Timing and bandwidth considerations}
\label{sec:timing_bandwidth}

Let $P$ denote the number of model parameters, $K$ the number of
replicas, and $N_{\mathrm{tok}}$ the global number of tokens processed
per inner step. Using the standard estimate of $6P$ FLOPs per token, the
duration of one inner step is
\[
T_{\mathrm{step}}
\simeq
\frac{6PN_{\mathrm{tok}}}
     {K\rho_{\mathrm{MFU}}F_{\mathrm{peak}}},
\qquad
T_{\mathrm{inner}}=HT_{\mathrm{step}},
\]
where $F_{\mathrm{peak}}$ is the peak accelerator throughput and
$\rho_{\mathrm{MFU}}$ is the model FLOPs utilization.

Assuming an FP32 outer gradient and a ring all-reduce, the
communication time is approximated by
\[
T_{\mathrm{comm}}
\simeq
\frac{2(K-1)}{K}
\frac{4P}{b_{\mathrm{net}}},
\]
where $b_{\mathrm{net}}$ is the effective inter-replica bandwidth.

The CPU--GPU transfer cost depends on the method. Let $c_m$ denote the
number of FP32 parameter-sized buffers transferred by method $m$ per
outer step. We use
\[
c_{\mathrm{DiLoCo}}
=
c_{\mathrm{Eager}}
=
c_{\mathrm{Delayed}}
=2,
\qquad
c_{\ours{}}=5,
\]
where the additional three buffers for \ours{} correspond to the
inactive AdamW state. The resulting outer-update cost is
\[
T_{\mathrm{outer}}^{(m)}
\simeq
T_{\mathrm{opt}}^{(m)}
+
c_m\frac{4P}{b_{\mathrm{CPU/GPU}}}.
\]
In the estimates below, we neglect the optimizer kernel time relative
to the transfer time.

The wall-clock time per communication round is therefore
\[
\begin{aligned}
T_{\mathrm{DiLoCo}}
    &= T_{\mathrm{inner}} + T_{\mathrm{comm}}
       + T_{\mathrm{outer}}^{(\mathrm{DiLoCo})},\\
T_{\mathrm{Eager}}
    &= \max\{T_{\mathrm{inner}},T_{\mathrm{comm}}\}
       + T_{\mathrm{outer}}^{(\mathrm{Eager})},\\
T_{\mathrm{Delayed}}
    &= \max\{T_{\mathrm{inner}},
             T_{\mathrm{comm}}
             +T_{\mathrm{outer}}^{(\mathrm{Delayed})}\},\\
T_{\ours{}}
    &= \max\{T_{\mathrm{inner}},
             T_{\mathrm{comm}}
             +T_{\mathrm{outer}}^{(\ours{})}\}.
\end{aligned}
\]

\paragraph{Representative H100 setting.}
For the $134$M model with $K=4$ and
$N_{\mathrm{tok}}=131{,}072$, we take
$F_{\mathrm{peak}}=1\,\mathrm{PFLOP/s}$,
$\rho_{\mathrm{MFU}}=0.5$,
$b_{\mathrm{net}}=10\,\mathrm{GB/s}$, and
$b_{\mathrm{CPU/GPU}}=3\,\mathrm{GB/s}$. This gives
\[
T_{\mathrm{step}}\simeq52.7\,\mathrm{ms},
\qquad
T_{\mathrm{comm}}\simeq80\,\mathrm{ms},
\qquad
\frac{4P}{b_{\mathrm{CPU/GPU}}}\simeq179\,\mathrm{ms}.
\]
For $H=30$, $T_{\mathrm{inner}}\simeq1.58\,\mathrm{s}$.

\begin{table}[t]
\centering
\caption{Timing estimates for the $134$M model under the representative
H100 setting described above. $H_{\mathrm{hide}}$ is the minimum number
of inner steps required to hide the complete outer path when the method
permits full overlap.}
\label{tab:timing-bandwidth}
\begin{tabular}{lcccc}
\toprule
Method
& $c_m$
& $T_{\mathrm{outer}}^{(m)}$
& $T_{\mathrm{round}}$ at $H=30$
& $H_{\mathrm{hide}}$ \\
\midrule
DiLoCo
& 2 & $0.36$ s & $2.02$ s & -- \\
Eager DiLoCo
& 2 & $0.36$ s & $1.94$ s & -- \\
Delayed DiLoCo
& 2 & $0.36$ s & $1.58$ s & 9 \\
\ours{}
& 5 & $0.89$ s & $1.58$ s & 19 \\
\bottomrule
\end{tabular}
\end{table}

For methods that overlap the complete outer path,
\[
H_{\mathrm{hide}}^{(m)}
=
\left\lceil
\frac{T_{\mathrm{comm}}+T_{\mathrm{outer}}^{(m)}}
     {T_{\mathrm{step}}}
\right\rceil.
\]
Thus, Delayed DiLoCo requires $H\geq9$, while the larger optimizer state
of \ours{} requires $H\geq19$. Both are below the $H=30$ used in our
main experiments, for which the outer path is fully hidden.

\section{\modif{Further Ablations}}
\label{sec:extra}

\subsection{\modif{Dynamics of the Parameter Distance \texorpdfstring{$\|\theta_t - \tilde{\theta}_t\|_2$}{||\theta_t - \tilde{\theta}_t||_2}}}
\label{sec:drift_ablation}

\noindent
\begin{minipage}[c]{0.55\textwidth}
\modif{To understand the interaction between the two parameter states throughout training, Figure~\ref{fig:l2_distance} tracks the $L_2$ distance $\|\theta_t - \tilde{\theta}_t\|_2$ over the whole training for $H$ ranging from $H=5$ to $H=75$ on the 134M model.}

\smallskip
\modif{The dynamics exhibit three distinct phases: \textbf{(1) Initialization:} both models start identical ($\theta_0 = \tilde{\theta}_0$); \textbf{(2) Transient drift:} early in training, the trajectories drift apart up to a peak around 0.35B tokens that scales monotonically with $H$; and \textbf{(3) Contraction:} as training progresses, the distance steadily contracts across all settings, effectively collapsing to zero at the end ($\|\theta_T - \tilde{\theta}_T\|_2 \approx 0$).}

\smallskip
\modif{Crucially, because $\theta_T \approx \tilde{\theta}_T$ at convergence, evaluating on $\theta_t$, $\tilde{\theta}_t$, or their pairwise average yields identical validation performance, validating our evaluation protocol.}
\end{minipage}\hfill
\begin{minipage}[c]{0.40\textwidth}
\centering
\includegraphics[width=\linewidth]{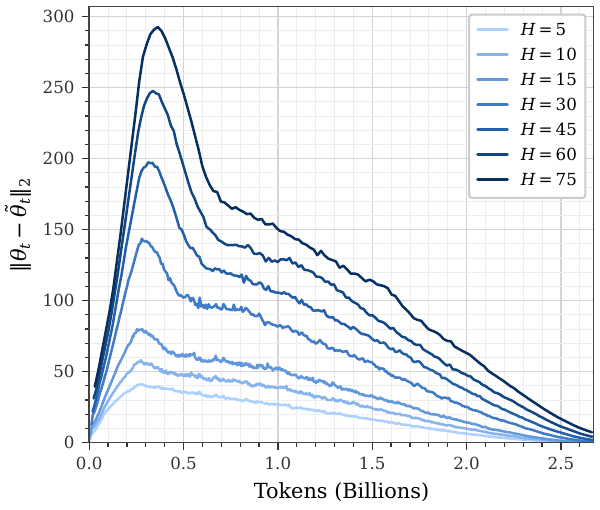}
\captionof{figure}{\modif{$L_2$ distance $\|\theta_t - \tilde{\theta}_t\|_2$ across training for varying $H$ on the 134M model.}}
\label{fig:l2_distance}
\end{minipage}

\subsection{\modif{Inner Optimizer State Policy}}
\label{sec:inner_optimizer_ablation}

\modif{In LACCO, each outer iteration alternates between a main stage and a lookahead stage. To evaluate how the inner AdamW state should be handled across these stages, we conduct an ablation on the 134M model with $K=4$ and $H=30$. 
\\
Let $S_0$ denote the inner optimizer state at the beginning of an iteration. We compare three policies: \textbf{(1) Rewind:} the lookahead stage initializes from $S_M$, but its state updates are discarded at the end of the round, restarting the next cycle from $S_M$; \textbf{(2) Shared state:} a single continuous AdamW state evolves sequentially through both stages ($S_0 \to S_M \to S_L$); and \textbf{(3) Momentum averaging:} both stages branch from $S_0$ to reach $S_M$ and $S_L$, and the next cycle initializes from $\frac{1}{2}(S_M + S_L)$.}

\begin{table}[H]
\centering
\small
\caption{\modif{Ablation of inner AdamW state policies on the 134M model at $K=4, H=30$.}}
\label{tab:inner_optimizer_ablation}
\begin{tabular}{lc}
\toprule
\modif{\textbf{Inner State Policy}} & \modif{\textbf{Val. Loss} $\downarrow$} \\
\midrule
\modif{LACCO, no averaging (rewind)} & \modif{3.029} \\
\modif{LACCO, no averaging (shared state)} & \modif{3.008} \\
\modif{\textbf{LACCO (momentum averaging, default)}} & \modif{\textbf{2.988}} \\
\bottomrule
\end{tabular}
\end{table}

\modif{As shown in Table~\ref{tab:inner_optimizer_ablation}, momentum averaging yields the best performance, achieving the lowest validation loss of 2.988. Maintaining a single shared state across both stages without averaging achieves 3.008, which still improves over the Eager DiLoCo baseline (3.020). In contrast, completely discarding the updates from the lookahead trajectory degrades the loss to 3.029.}

\subsection{\modif{Peak Memory Footprint and Training Throughput}}
\label{sec:memory_throughput_ablation}

\modif{To evaluate the practical overhead of maintaining two sets of parameter and optimizer states, Table~\ref{tab:memory_throughput} reports the peak allocated GPU memory and end-to-end token throughput for $K=4$ workers training the 0.5B model at $H=30$ on 80GB NVIDIA A100 GPUs across inter-node bandwidths of 0.5, 1, and 5\,Gbps.}

\begin{table}[H]
\centering
\small
\caption{\modif{Peak GPU memory allocation and measured training throughput across bandwidths for the 0.5B model ($K=4, H=30$). Percentages indicate memory utilization relative to the 80\,GB GPU capacity.}}
\label{tab:memory_throughput}
\begin{tabular}{lcccc}
\toprule
\modif{Method} & \modif{Peak GPU memory} & \modif{0.5 Gbps} & \modif{1 Gbps} & \modif{5 Gbps} \\
\midrule
\modif{DiLoCo}             & \modif{37.9 GiB (48\%)} & \modif{77.7k tok/s}  & \modif{103.0k tok/s} & \modif{138.8k tok/s} \\
\modif{LACCO, CPU offload} & \modif{40.3 GiB (51\%)} & \modif{145.5k tok/s} & \modif{147.8k tok/s} & \modif{148.4k tok/s} \\
\modif{LACCO, fully GPU}   & \modif{56.1 GiB (71\%)} & \modif{146.7k tok/s} & \modif{148.2k tok/s} & \modif{148.5k tok/s} \\
\bottomrule
\end{tabular}
\end{table}

\modif{These results highlight three main findings: (1) Minimal memory overhead with offloading: when offloading inactive optimizer states to the host CPU, peak GPU memory increases by only 2.4\,GiB over synchronous DiLoCo (40.3\,GiB vs.\ 37.9\,GiB), making it practical even under tight memory budgets. (2) Resilient throughput: LACCO sustains a nearly constant throughput of $\sim$146k--148.5k tok/s across all network conditions, whereas DiLoCo's throughput drops by nearly half (to 77.7k tok/s) at 0.5\,Gbps due to blocking synchronization; and (3) Zero throughput penalty from offloading: the throughput difference between CPU-offloaded and fully GPU-resident LACCO is practically nonexistent (e.g., 147.8k vs.\ 148.2k tok/s at 1\,Gbps), confirming that host transfers are entirely hidden behind local computation.}

\end{document}